\documentclass[letterpaper, 10 pt, conference]{ieeeconf}  

\IEEEoverridecommandlockouts                              

\usepackage{amsmath,amsfonts,amssymb,mathrsfs}
\usepackage{graphicx}
\usepackage{psfrag,graphicx,epsfig,epsf}
\usepackage[ruled,linesnumbered, noend]{algorithm2e}
\usepackage{fancyhdr}
\usepackage{wrapfig}
\usepackage{subfigure}
\usepackage{mathtools}
\usepackage{url}
\usepackage{color}
\usepackage{verbatim}
\usepackage{xspace}
\usepackage{cite} 
\usepackage{wrapfig}
\usepackage{numprint}
\usepackage{placeins}
\usepackage[noend]{algpseudocode}

\usepackage{xspace}
\usepackage{xcolor}
\definecolor{gold}{HTML}{BE830E}
\usepackage[pdfa,colorlinks,bookmarksopen,bookmarksnumbered,allcolors=gold]{hyperref}

\usepackage{verbatim}
\usepackage{subcaption}
\usepackage{booktabs}
\usepackage{multirow}
\usepackage{multicol}
\usepackage{tabularx}
\usepackage{dsfont}
\usepackage{lipsum}
\usepackage[table]{xcolor}

\usepackage{pgfplots}
\pgfplotsset{compat=1.18}
\usetikzlibrary{calc}

\title{\LARGE \bf
Virtues of Ordered Chaos:\\ Planning with Topple Actions 
in Tabletop Stack Rearrangement
}

\author{Hao Lu and Rahul Shome
\thanks{*The authors are affiliated to the School of Computing at the Australian
National University, Canberra, AU. Email: hao.lu@anu.edu.au}
}

\newcommand{\rahul}[1]{{#1}}

\begin{document}
\newtheorem{theorem}{Theorem}
\newtheorem{definition}{Definition}
\newtheorem{lemma}{Lemma}
\newtheorem{corollary}{Corollary}
\newtheorem{assumption}{Assumption}
\newtheorem{condition}{Condition}
\newtheorem{remark}{Remark}
\newtheorem{claim}{Claim}

\newcommand{\real}{\mathbb{R}}
\newcommand{\cspace}{\mathcal{X}}
\newcommand{\state}{x}
\newcommand{\cfree}{\cspace_{\mathrm{free}}}
\newcommand{\cobs}{\cspace_{\mathrm{obs}}}
\newcommand{\startstate}{\state_{s}}
\newcommand{\goalstate}{\state_{g}}
\newcommand{\traj}{\pi}
\newcommand{\trajset}{\Pi}
\newcommand{\cost}{c}
\newcommand{\opt}[1]{#1^*}
\newcommand{\opttraj}{\opt{\traj}}
\newcommand{\optcost}{\opt{\cost}}
\newcommand{\n}{n}
\renewcommand{\d}{d}
\newcommand{\reald}{\real^\d}
\newcommand{\wspace}{\mathcal{W}}
\newcommand{\robot}{r}
\newcommand{\wobs}{\wspace_{\mathrm{obs}}}
\newcommand{\wrob}{\wspace_{\mathrm{\robot}}}
\newcommand{\vol}{{vol}}
\newcommand{\dist}[1]{\pmb {#1}}
\newcommand{\wsample}{w}
\newcommand{\wsubset}{\mathbbm{\mathfrak{W}}}
\newcommand{\wsubsubset}{\mathbbm{\mathcal{F}}}
\newcommand{\wevents}{\wsubsubset^\wspace}
\newcommand{\probspace}{\mathbb{P}}
\newcommand{\col}{\mathbbm{1}_{\mathrm{col}}}
\newcommand{\pcol}{\prob_{\mathrm{col}}}
\newcommand{\wrand}{\mathrm{W}}
\newcommand{\NUM}{N}

\newcommand{\unitd}{[0,1]^\d}
\newcommand{\rn}{r_\n}
\newcommand{\sn}{s_\n}
\newcommand{\rlogn}{\Big( \frac{\log \n}{\n} \Big)^{ \frac{1}{\d} }}
\newcommand{\fn}{f_{\n}}
\newcommand{\rfn}{\Big( \frac{\fn}{\n} \Big)^{ \frac{1}{\d} }}
\newcommand{\graph}{\mathcal{G}_\n}
\newcommand{\nodes}{\mathcal{V}_\n}
\newcommand{\edges}{\mathcal{E}_\n}
\newcommand{\clearance}{\delta}
\newcommand{\ball}[2]{\mathcal{B}_{#2}(#1)}
\newcommand{\prob}{\mathrm{Pr}}
\newcommand{\ntends}{\n\rightarrow\infty}
\newcommand{\lattice}{\mathcal{L}}
\newcommand{\length}{\ell}
\newcommand{\width}{w}
\newcommand{\sepl}{s_\length}
\newcommand{\sepw}{s_\width}
\newcommand{\sitep}{p^\lattice}
\newcommand{\sitepn}{p^\lattice_\n}
\newcommand{\tcite}[1]{{\color{red}[#1]}}
\newcommand{\directed}{\overrightarrow{\lattice}}
\newcommand{\dual}{{\underline{\lattice}}}
\newcommand{\pathlattice}{\directed^\traj}
\newcommand{\crossing}{\directed_{\times}}
\newcommand{\dualcrossing}{\dual_{\times}}
\newcommand{\mbold}[1]{{#1}}
\newcommand{\CI}{K}
\newcommand{\CII}{{C_1}}
\newcommand{\CIII}{{C_2}}
\newcommand{\CIV}{C_3}
\newcommand{\CV}{C_4}
\newcommand{\CVI}{C_5}
\newcommand{\CVII}{C_6}
\newcommand{\CVIII}{C_7}
\newcommand{\CVIX}{C_8}
\newcommand{\CVX}{C_9}
\newcommand{\posevent}{\mathfrak{E}}
\newcommand{\vect}{\vec{v}}
\newcommand{\uect}{\vec{u}}


\newcommand{\obj}{o}
\newcommand{\objects}{\mathcal{O}}
\newcommand{\scene}{S}
\newcommand{\scenes}{\mathcal{S}}
\newcommand{\rand}[1]{\mathbf{#1}}
\newcommand{\samples}{k}
\newcommand{\image}{\mathcal{I}}
\newcommand{\frust}{{F}}
\newcommand{\scenevolume}{{V}}
\newcommand{\volume}{\mathcal{V}}
\newcommand{\genai}{\mathcal{M}}
\newcommand{\objsem}{object\xspace}
\newcommand{\ssh}{\mathcal{D}}
\newcommand{\pr}{\mathbb{P}}
\newcommand{\event}{\mathcal{E}}
\newcommand{\bb}{\mathbb{B}_o}
\newcommand{\ssd}{\mathcal{D}_{\image}}
\newcommand{\sd}{\mathcal{D}_{\image\obj}}
\newcommand{\sh}{\tilde{\mathcal{D}}_{\image\obj}}
\newcommand{\inim}{\mathcal{I}}
\newcommand{\gtd}{\Gamma_{\image}}
\newcommand{\gtdo}{\Gamma_{\image\obj}}
\newcommand{\conf}{\omega_{\obj}}
\newcommand{\outim}{\mathbb{I}}
\newcommand{\outvol}{\mathbb{V}}
\newcommand{\e}{\mathbf{x}}
\newcommand{\gspan}{\mathcal{X}}
\newcommand{\ent}{\mathcal{H}}
\newcommand{\objcond}[1]{#1_\obj}
\newcommand{\cent}{\mathcal{H}^{\times}}
\newcommand{\gtsd}{\mathcal{G}_{\inim\obj}}
\newcommand{\ggrid}{\tilde{\gtsd}}
\newcommand{\dnn}{\Delta}
\newcommand{\rwa}{\mathcal{A}}
\newcommand{\ltask}{\task}
\newcommand{\expect}{\mathbb{E}}
\newcommand{\falseneg}{\text{FNR}}
\newcommand{\indi}{\mathds{1}}
\newcommand{\nyu}{NYU Depth V2\xspace}
\newcommand{\ct}{\vspace{-1em}}
\newcommand{\rk}{RealEstate10k\xspace}
\newcommand{\gtlabel}{y_{\inim\obj}}
\newcommand{\plabel}{\hat{y}_{\inim\obj}}
\newcommand{\diam}{\mathtt{diam}}


\newcommand{\matterport}{Matterport3D}

\newcommand{\object}{o}
\newcommand{\wobj}{\wspace_{\mathrm{\object}}}
\newcommand{\spsem}[1]{\Phi_{\mathrm{#1}}}
\newcommand{\realiz}{\wspace^i}
\newcommand{\consem}[1]{\mathbbm{1}_{\mathrm{#1}}}

\newcommand{\envsample}[1]{\sim}
\newcommand{\genmo}{\mathbf{g}_\theta}
\newcommand{\sampler}{\mathbf{s}}
\newcommand{\observ}{O}
\newcommand{\gprobspace}{\bar{\probspace}}

\newcommand{\mytexttilde}{\raisebox{0.5ex}{\texttildelow}}


\newcommand{\objset}{\mathcal{O}}
\newcommand{\pose}{l}
\newcommand{\posespace}{P}
\newcommand{\taskspace}{\mathcal{T}}
\newcommand{\actionset}{\mathcal{A}}
\newcommand{\action}{a}
\newcommand{\tables}{\mathtt{T}}

\maketitle
\thispagestyle{empty}
\pagestyle{empty}

\begin{abstract}

Efficient object manipulation strategies have significant impact in automation applications. In this work, the stack rearrangement in tabletop settings is studied, with a focus on augmenting the task planning domain with richer non-prehensile aggregating actions, in particular the toppling of objects from a stack to the table. Toppling can compress long sequences of intermediate relocations. Computed plans need to interleave pick-and-place actions with topple throughout its plan based on the problem.  In order to generate the task plan and model an abstraction to compute solutions that include both pick-and-place and topple actions, a novel aggregating gadget for topple is introduced. Using this directed graphical abstraction, candidate task plan computation becomes a variant of the pebble motion problem, treating objects as pebbles. Benchmarks are then reported in a IsaacSim-based physics simulation. Results highlight clear benefits of achieving faster execution than solely using pick-and-place actions. Though this work primarily investigates the topple action, we demonstrate that similar abstractions can model other aggregating actions of interest, like scoop. The current work provides a preliminary, strong indication of the promising benefits of abstractions for rich object interactions in manipulation applications. 
\end{abstract}

\section{Introduction}
Tabletop manipulation problems arise in a wide variety of automation applications. Improving the efficiency of object rearrangement in tabletop settings has been well studied and represents a key enabling line of research for logistics, warehousing, and product sorting.
In this work we focus on \emph{object stack rearrangement}, where objects occupy a finite set of stacking locations and must be moved from an initial stacked arrangement to a target arrangement.

\begin{figure}[t]
    \centering
    \includegraphics[trim=10cm 0 10cm 0,clip,width=0.49\linewidth]{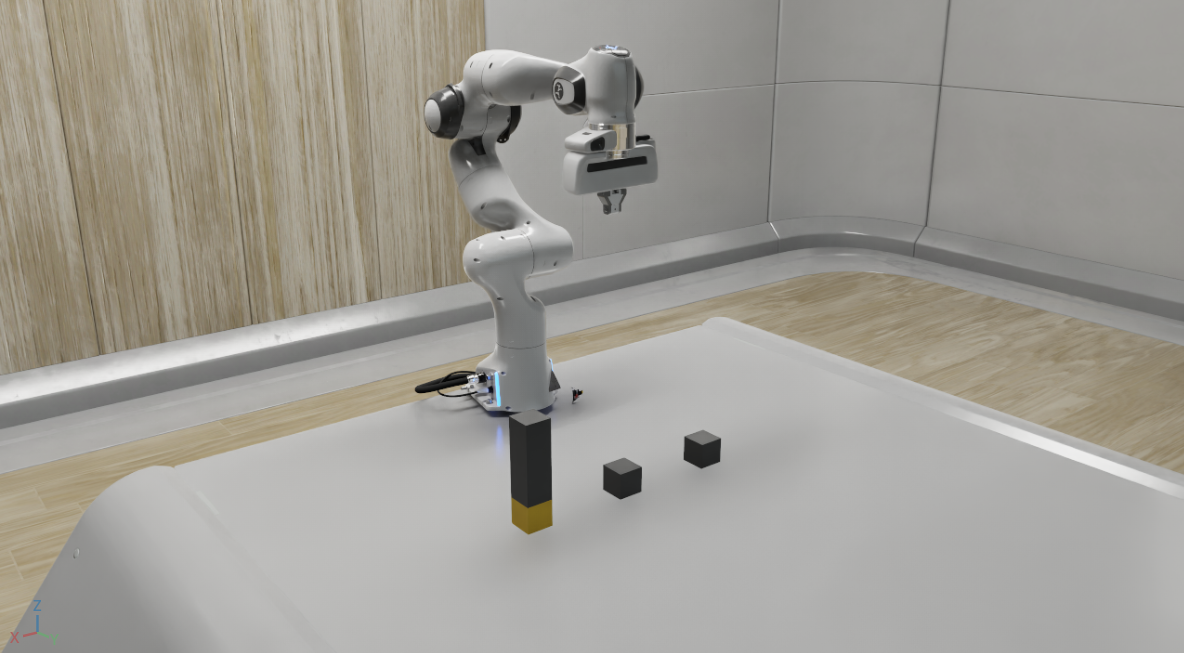}
    \includegraphics[trim=10cm 0 10cm 0,clip,width=0.49\linewidth]{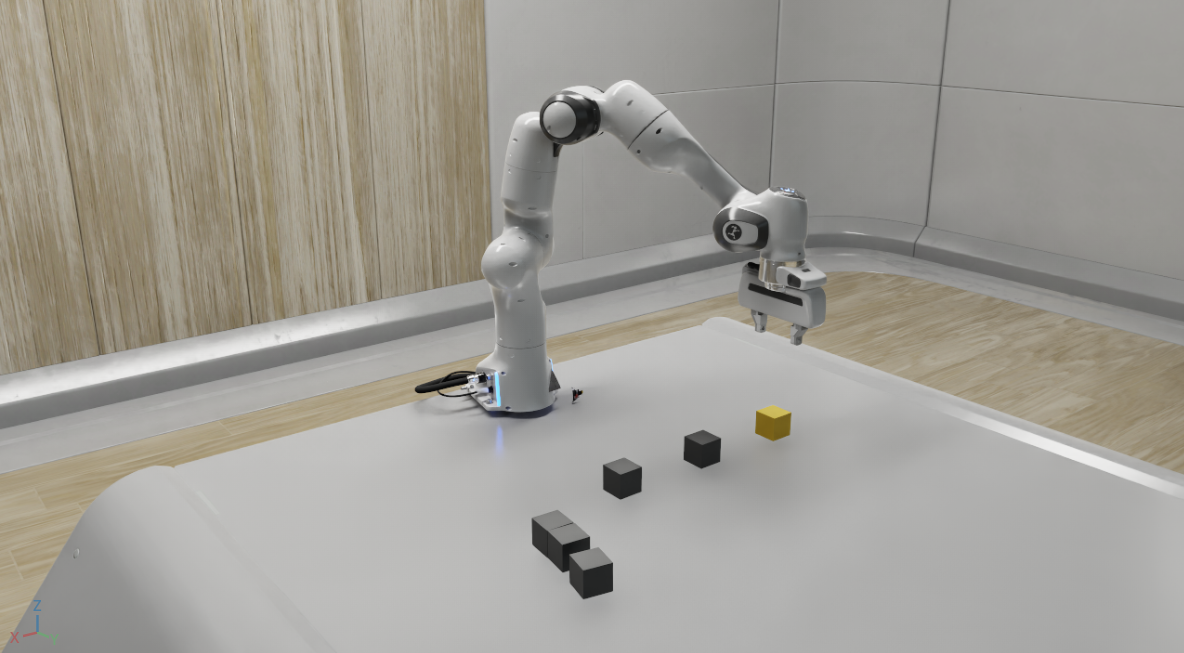}
    \caption{Certain stack rearrangement tasks can be made significantly more efficient using topple actions. Here, the highlighted object must be moved to the target location. With pick--and--place only, the blocking objects must be relocated one-by-one (8 actions total). A single topple clears multiple blockers simultaneously, solving the instance in 3 actions.}
    \label{fig:fig1}
    \vspace{-0.25in}
\end{figure}

Most prior stack rearrangement planners~\cite{han2017complexity,han2018efficient} consider prehensile \emph{pick--and--place} actions. In contrast, this work focuses on efficiency gains achievable from non-prehensile aggregating object interactions in tabletop stack rearrangement. Pushing is a classic example~\cite{dogar2011framework,huang2019large,vieira2022persistent} of non-prehensile manipulation. 
We introduce and primarily study \emph{topple} action: by deliberately toppling a stack segment, multiple blocks can fall to the table at once, exposing them to make subsequent picks feasible. This potentially eliminates long sequences of intermediate picks and places (Fig.~\ref{fig:fig1}) from the solution. 
The contact mechanics of toppling primitives has been studied~\cite{lynch1999toppling} though its integration in multi-object rearrangement is underexplored. Notably, toppling objects indiscriminately might not improve the solution in general.
Deciding which objects to topple and at what juncture of the plan, that can involve a sequence of multiple interleaved pick-and-place actions with topple, is still under the purview of planning. \textit{A planning abstraction} that admits such solutions is critical alongside a co-evolved execution strategy. Our focus is on representing topple {as an aggregating action} acting simultaeneously on multiple objects within stack rearrangement planning and partially grounded execution.

The dynamic nature of toppling benefits us but introduces a key planning difficulty: after a topple, the exact poses of toppled objects are not known \emph{a priori} to the planner. Classical task and motion planning~\cite{dantam2018incremental} typically assumes grounded object poses when generating downstream actions. Rather than predicting or enumerating topple outcomes, we aim to \emph{plan and execute} topple within a task-and-motion pipeline by leaving topple effects \emph{partially grounded} and grounding only what is needed at execution time, similar to recent work~\cite{pan2024task}. It remains to be seen how such a task plan interleaving prehensile and such motivating non-prehensile aggregating actions can be computed efficiently, addressing which forms a key focus of this work. The partially grounded nature of toppling still conveniently expresses natural symbolic conditions, for instance, \textit{toppled objects are likely to rest on the tabletop, albeit in unknown poses, and can be picked}. Using appropriate abstractions can capture the partially grounded preconditions and effects of such actions and will still be beneficial to devise candidate plans. If the candidate plan is successful, such an abstraction should admit strictly better solutions at the task level, since the nominal pick-and-place solution is part of the abstraction. It is an open question whether these candidate plans are indeed helpful when executed and succeed at speeding up stack rearrangement. The current work shows that this is indeed the case, under certain conditions. Note that the category of stack problems problems with topple was erstwhile unsolved. The proposed analysis of task-level abstractions and partially grounded planning strategies with rigorously characterized conditions of feasibility and efficiency is a necessary push in the direction towards exploring \rahul{a richer category of manipulation actions in object rearrangement.} 


\noindent\textbf{Contributions.}
This paper makes the following contributions:
(i) we introduce the \emph{SMASH gadget}, a task-level graphical abstraction that represents \emph{topple} as a simultaneous multi-object transition via capacity constraints.
(ii) We design a \emph{mode-graph} abstraction that supports computing candidate task plans which \emph{interleave} topple and pick--and--place actions for stack rearrangement, along with a statement of the assumptions and conditions under which the abstraction admits feasible plans and efficient solutions, and the discussion of the trade-offs introduced by using uncertain actions like topple based on empirical observations.

We validate our contributions through simulated benchmarks on IsaacSim, demonstrating that the resulting plans with topple can be executed 
by integrating them into a partially grounded TAMP framework.
Across benchmarks, the introduction of the topple action reduces execution times by effectively “compressing” long pick–and–place sequences into single multi-object transitions.
Benefits are further reported in motivating preliminary demonstration that repurposes a similar gadget on a different aggregating action (scoop) to simultaneously transport objects. This suggests that the proposed graphical abstraction principles can extend beyond toppling to other simultaneous aggregating non-prehensile actions. The current study lays the necessary building block towards incorporating broader and richer object interactions and future investigation of practical considerations for real-robot considerations.

\section{Related Work}
\noindent\textbf{Object Rearrangement}: In manipulation, object rearrangement~\cite{ota2004rearrangement} is a category of task and motion planning problems involving interactions between objects and manipulators with the goal of changing object locations. 
Theoretical underpinnings have been characterized through efficient solutions for variants involving only one object pick-and-place (monotone)~\cite{stilman2007manipulation}. 
Generalized reductions to related computational problems assist in deriving efficient solutions~\cite{han2017complexity}. 
Closely related to this work is the problem of rearranging objects in stacks~\cite{han2018efficient}, where task-specific gadget abstractions capture key combinatorial structure. Multi-arm applications were explored by leveraging underlying problem structure~\cite{shome2020synchronized}. 
The current paper follows this line of work in identifying efficient abstractions to guide task planning in object rearrangement, with the distinction that here the focus is on stack rearrangement when the action set is augmented with a non-prehensile \emph{topple} primitive.


\begin{figure*}[t]
    \centering
    \includegraphics[height=2.5in]{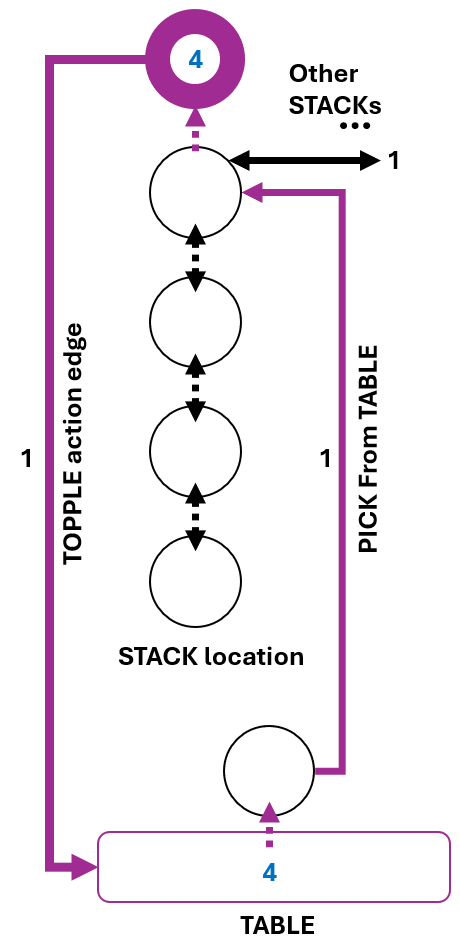}
    \includegraphics[height=2.5in]{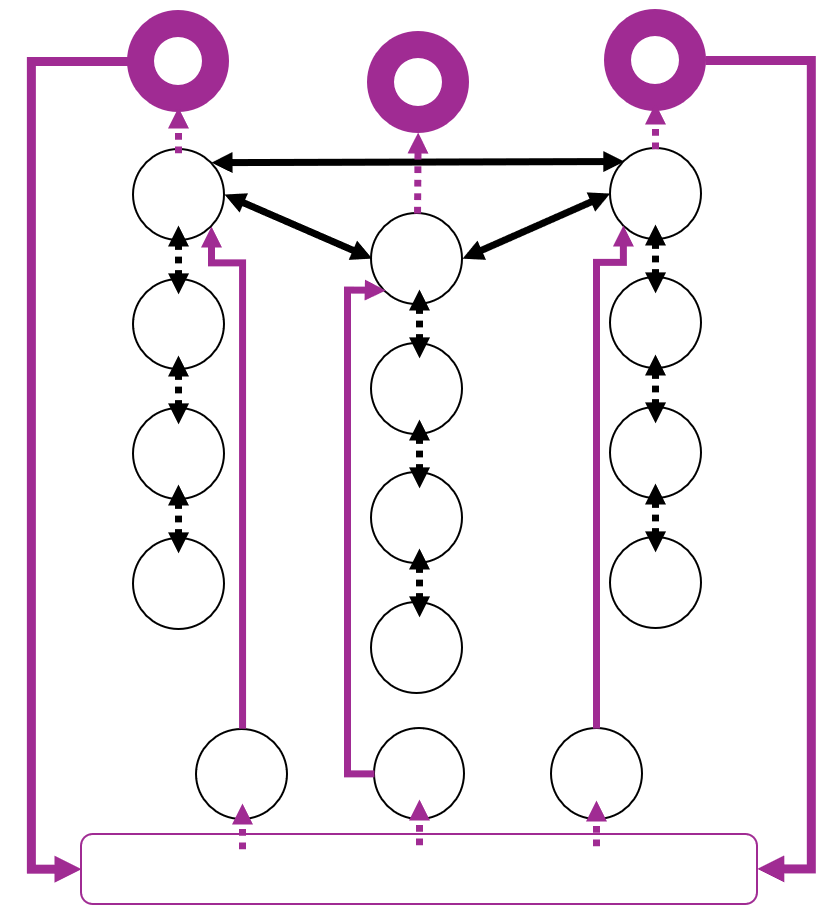}
    \includegraphics[height=2.5in]{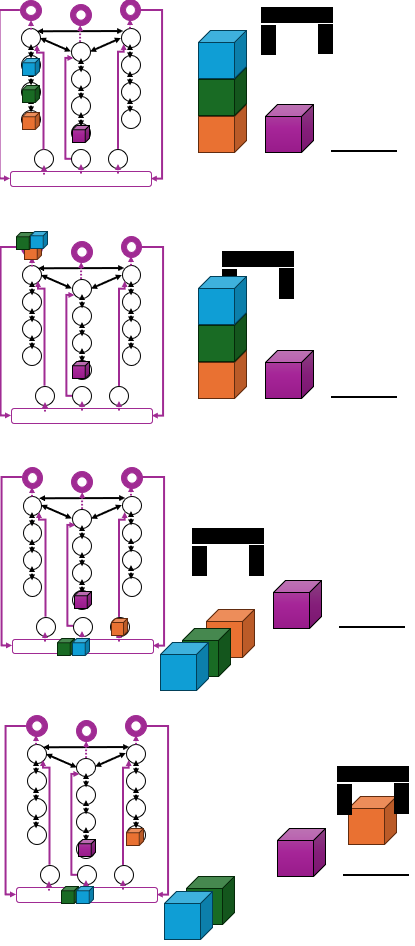}
    \caption{The SMASH gadget over which the object rearrangement problem is modeled as a modified pebble motion problem over the directed graph with capacity constraints. The pink edges are involved with topple actions. 
    The figure in the middle denotes a problem comprising three STACK locations of maximum height 4. 
    The right figure illustrates a solution to the pebble motion problem for four objects where the goal is for the bottom orange object to be moved to the empty STACK location. TOPPLE serves as an aggregate action that allows more efficient task plans with fewer actions.}
    \label{fig:gadgets}
    \vspace{-0.2in}
\end{figure*}

\noindent\textbf{Non-Prehensile Manipulation:} 
Seminal work~\cite{dogar2011framework} focused on non-prehensile push actions within the context of manipulation. Recent work~\cite{vieira2022persistent} has explored topological information to guide non-prehensile interactions with groups of objects. Pushing can assist in precise and robust execution in some cases~\cite{shome2019towards}. Often it is recognized as more efficient to apply large-scale actions that can move multiple objects with a single interaction~\cite{huang2019large}. 
In the task planning layer, push and pick-and-place should be interleaved seamlessly in the task~\cite{pan2021decision}. Thus far, most task-planning treatments of non-prehensile actions have focused on pushing variants. In this work, we focus on \emph{topple}. 

Toppling, despite being a natural and common affordance, is relatively underexplored as a planning primitive. Seminal study~\cite{lynch1999toppling} characterized fundamental features of this interaction. Our approach does not rely on predicting topple outcomes. Instead, it uses a task-level abstraction that treats topple as a simultaneous multi-object transition, and leverages partially grounded execution~\cite{pan2024task} to handle downstream motions whose feasibility depends on post-topple poses.

\noindent\textbf{Task and Motion Planning for Manipulation: }
Sequential manipulation presents itself as a task and motion planning problem~\cite{garrett2021review}. 
Various hierarchical strategies have been proposed through literature~\cite{akbari2018combined,garrett2018ffrob,dantam2018incremental,Kaelbling:2011gb} over actions and motions.
A typical abstraction to handle task planning~\cite{dantam2018incremental} is to use satisfiability modulo theories (SMT) solvers commonly defined over PDDL specifications. While PDDL is a general-purpose domain modeling tool, the contrast here is to design a task-specific abstraction that captures the combinatorial structure of stacking with toppling.

The uncertainty of non-prehensile interaction draws parallels to task and motion planning variants that delve into stochastic domains~\cite{shah2020anytime,pan2022failure}, assuming a known transition model.
Some work has started interleaving execution in incremental replanning~\cite{curtis2022M0M} to update models. 
Recent work has introduced partially grounded task and motion planning and execution~\cite{pan2024task}. 
This partially grounded paradigm is leveraged in the current work for modeling the topple action, balancing motion planning offline with a reliance on computed action groundings during execution.

\noindent\textbf{Learning: }
Learned and dexterous manipulation motions of the end effector recovered from demonstrations~\cite{chi2023diffusion} can be complementary to the categories of non-prehensile actions being described in the current work. Some characterization of the preconditions and effects of learned controllers~\cite{vieira2024morals}. 
In the current work, the focus is on theoretically devised combinatorial abstractions that are interpretable and supplement our understanding of the problem domain, potentially enabling efficient expert training data and complementary advances in learning in the future.

\section{Problem Formulation}
\label{sec:problem}
\subsection{Preliminaries}

Consider a manipulator in a workspace $\wspace\subset\mathbb{R}^3$, with $\d$ degrees of freedom, describing a configuration in a configuration space $\state \in\cspace\subset\reald$, its collision subset $\cobs$ and feasible subset $\cfree = \cspace \setminus \cobs$. There are $\ell$ objects $\objset = \{\object^1,\cdots\object^\ell\}$ with poses $(\pose^1,\cdots,\pose^\ell)$. For manipulation planning, the fully defined combined planning space is the combined state $(\state,\pose^1\cdots\pose^\ell)$. 

The discrete planning domain is defined by symbolic actions $\action \in \actionset$ associated with the combined state. Each action $\action$ is associated with a feasible manipulator motion $\traj:[0:1] \rightarrow \cfree$. This is an action-motion pair $(\action,\traj)$ that constitutes a step of a longer task and motion plan. 

The tabletop stack domain is restricted to states where each object pose $\pose^i$ corresponds to either resting on the surface of the table $\tables$ or being supported by another object $\object^j$ at pose $\pose^j$. We assume a finite set of $S$ reachable stack locations on the table (including buffers), consistent with gadget-based stack rearrangement abstractions used later. Objects may be moved between these stack locations via pick-and-place, and may be moved from stacks to the table via topple.

\begin{definition}[Tabletop Object Stack Rearrangement]
The object rearrangement problem is defined for a manipulator and $\ell$ objects at an initial state $(\startstate, \pose^1_s \cdots \pose^\ell_s )$ and a target set of object poses $(\pose^1_g\cdots \pose^\ell_g)$. A solution generates a task and motion plan $\Pi$ as a sequence of actions and continuous arm motions $\Pi = \{(\action^1,\traj^1) \cdots (\action^H,\traj^H)\}$, such that the application of this plan results in a final state $(\goalstate, \pose^1_g \cdots \pose^\ell_g )$.  
\end{definition}

Note that while the discussion and the method will be framed for a single goal, the extension to goal sets is straightforward and will be omitted for brevity.

\subsection{Partial grounding and topple}

A fully grounded step $(\action^h, \traj^h)$ is one where the state before its application $(\state^h,\pose^{1,h}\cdots\pose^{\ell,h})$ is known and the effect-state can also be grounded $(\state^{h+1},\pose^{1,h+1}\cdots\pose^{\ell,h+1})$. A partially grounded action is one where these might not be known, might not be analytically describable, or easy to simulate. Such steps will be assumed to leave corresponding entries empty in the symbolic state, e.g., $(\state^h,\pose^{1,h}\cdots \phi^{\cdot,h}\cdots \pose^{\ell,h})$.

The topple action applied at step $h$ to $m$ objects $\object^k\cdots \object^{k+m}$ is a partially grounded action where the start state can be fully known with poses $\pose^k\cdots \pose^{k+m}$. However, after the application of the action, the state will be partially grounded as $(\state^{h+1},\pose^{1,h+1} \cdots \phi^{k,h+1} \cdots \phi^{k+m,h+1} \cdots \pose^{\ell,h+1})$. This arises because the exact post-topple poses are not known apriori and can only be observed during execution.

A partially grounded task and motion plan~\cite{pan2024task} for tabletop object stack rearrangement can therefore include topple steps (and subsequent steps that depend on topple outcomes) whose geometric details are deferred and grounded online during execution.

\subsection{Foundations}

As in prior graph-based formulations of rearrangement planning~\cite{shome2020synchronized}, certain structural conditions are required to ensure that objects located in designated regions can be manipulated independently. 

\begin{condition}[Object Non-interactivity of Stack Locations]
In an object stack rearrangement problem with $S$ reachable stack locations, the feasibility of the pick action for an object at the top of one stack does not depend upon the locations of other objects.
\end{condition}
This condition ensures that stack locations (including designated buffers) can be treated as independently accessible at the task level. In practice, stack and buffer locations are selected or computed to satisfy this separability property. As a result, candidate task plans discovered in the abstraction can be executed by appropriately ordering pick operations.
Introducing topple requires extending this structural reasoning to actions that simultaneously affect multiple objects. To formalize this extension, we define the following operator.

\begin{definition}[Monotone Transformation Operator]
An action $a$ associated with objects $\object^k, \ldots, \object^{k+m}$ and a workspace region $\tables_a \subseteq \tables$ is a \emph{monotone transformation operator} if, after applying $a$, relocating the affected objects to reachable target locations disjoint from $\tables_a$ yields a monotone rearrangement problem.
\end{definition}


Intuitively, a monotone transformation operator moves a set of objects into a designated intermediate region without introducing cyclic dependencies that would prevent subsequent sequential manipulation.
A properly defined topple action satisfies this definition: after toppling, the abstraction represents the affected objects as residing within a designated tabletop region $\tables_a$, and subsequent manipulation consists of pick-and-place operations that transfer objects from $\tables_a$ to stack locations.
Existing conditions for tabletop rearrangement~\cite{han2017complexity} already typically require a monotone property to hold in order to allow a sequence of picks in arbitrary order. This is combined with topple via the following condition.


\begin{condition}[Monotone Transformation Non-interactivity]
After applying a monotone transformation operator to $m$ objects, the feasibility of picking one object from $\tables_a$ does not depend on the arrangement of other objects within $\tables_a$.
\end{condition}




This condition justifies abstracting toppled objects at the task level as a set residing in $\tables_a$ without encoding their exact geometric configuration. Under this abstraction, individual picks from $\tables_a$ to stack locations can be scheduled independently at the symbolic level. The designated region $\tables_a$ is chosen in the planning abstraction to be disjoint from stack locations, preserving structural separability. Precise landing poses are resolved at execution time.

Under these conditions, topple can be represented symbolically as a simultaneous transition that moves multiple objects from stack locations into $\tables_a$ in a partially grounded manner. Subsequent pick-and-place actions from $\tables_a$ to stack locations are enabled in the abstraction, with their geometric realization deferred until execution. This structure forms the basis for the mode-graph abstraction developed in the next section, which enables computation of task plans that interleave topple and pick-and-place actions.



\section{Method}
This section introduces the task-planning abstraction that incorporates topple as a simultaneous multi-object action, and composing a partially grounded TAMP framework. 

\subsection{Gadget-based abstraction for stack rearrangement}
\begin{wrapfigure}{r}{0.3\linewidth}
    \centering
    \includegraphics[width=0.95\linewidth]{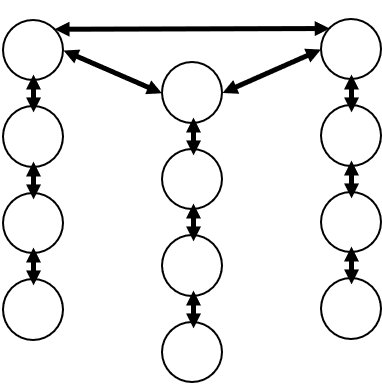}
    \caption{Pick and place gadget with 3 stacks. }
    \label{fig:papgadget}
\end{wrapfigure}

Prior work in object stack rearrangement~\cite{han2018efficient} introduced gadget constructions in which a pick--and--place stack rearrangement instance can be represented as moving $\ell$ object ``pebbles'' on a directed graph. We use a 
variant with equivalent connectivity properties (Fig.~\ref{fig:papgadget}): each stack is a chain of height-indexed locations, and the tops of stacks connect to enable pick--and--place transfers between stacks.

\begin{definition}[Pick and place gadget]
A pick and place gadget for $S$ stack locations and $\ell$ objects is a directed graph with node capacities and edge costs/capacities:
\vspace{-0.05in}
\begin{align*}
    G_{PAP} = &(V_{PAP}, E_{PAP})\\
    V_{PAP} = &\{(v_s^i, 1)\;|\; i \in [1\cdots \ell],\; s \in [1\cdots S]\}\\
    E_{PAP} = &\{(v_s^i, v_s^{i+1}, 0, 1)\;|\; i\in[1\cdots \ell\!-\!1],\; s\in[1\cdots S]\}\ \cup\\
              &\{(v_u^\ell, v_w^\ell, 1, 1)\;|\; u\neq w \}.
\end{align*}
Each node has capacity 1. Within-stack edges have cost 0, and cross-stack edges between stack tops have cost 1, corresponding to one pick--and--place action.
\end{definition}

In $G_{PAP}$, the tabletop is implicit: objects are always located at some stack location (including buffers). Introducing topple requires representing (i) a simultaneous transition that moves multiple objects off stacks, and (ii) subsequent picks for objects whose post-topple poses are not grounded at planning time.

\subsection{SMASH gadget}

\noindent\textbf{Simultaneous Manipulation Actions for Sequential Heuristics (SMASH).}
We introduce additional nodes and edges to represent topple and subsequent pick actions via a tabletop abstraction $\tables_a$. Importantly, we do not attempt to predict exact post-topple poses at planning time. Instead, toppled objects are treated as being on the tabletop (represented by $\tables_a$), and their precise poses are grounded during execution as needed.

For each stack we introduce a \emph{TOPPLE staging node} that can temporarily hold multiple objects (capacity $\ell$), a single \emph{TABLE node} representing $\tables_a$ (capacity $\ell$), and $\ell$ unit-capacity \emph{table pick} nodes that act as an interface for picking individual objects from the table back onto stack tops.

\begin{definition}[SMASH Gadget]
The SMASH gadget is defined as $G_{SMASH}=G_{PAP}\cup G_{SMASH}^+$ where:
\begin{align*}
    G_{SMASH}^+ = &(V_{SMASH}^+, E_{SMASH}^+)\\
    V_{SMASH}^+ = &\{(v_{T_s}, \ell)\;|\; s \in [1\cdots S]\}\ \cup\\
    &\{(v_{\tables},\ell)\}\ \cup\\
    &\{(v_{\tables}^i,1)\;|\; i\in[1\cdots \ell]\}\\
    E_{SMASH}^+ = &\{(v_s^\ell, v_{T_s}, 0, 1)\;|\; s\in[1\cdots S]\}\ \cup\\
                  &\{(v_{T_s}, v_{\tables}, 1, \ell)\;|\; s\in[1\cdots S]\}\ \cup\\
                  &\{(v_{\tables}, v_{\tables}^i, 0, 1)\;|\; i\in[1\cdots \ell]\}\ \cup\\
                  &\{(v_{\tables}^i, v_s^\ell, 1, 1)\;|\; i\in[1\cdots \ell],\ s\in[1\cdots S]\}.
\end{align*}
Here $v_{T_s}$ is the TOPPLE node for stack $s$, $v_{\tables}$ is the TABLE node representing $\tables_a$, and $v_{\tables}^i$ are unit-capacity interface nodes for picking from the table.
Edges $(v_s^\ell, v_{T_s})$ select objects to be toppled (cost 0, capacity 1 per object), edge $(v_{T_s}, v_{\tables})$ represents the topple action itself (cost 1, capacity $\ell$ to allow a simultaneous multi-object transition), and edges from the table interface nodes back to stack tops represent pick--and--place actions for toppled objects.
\end{definition}

The TABLE node deliberately abstracts away ordering and exact poses: once objects enter $v_{\tables}$, the planner only records that they reside somewhere on the tabletop (in $\tables_a$) until execution-time sensing grounds a specific pick pose. Under the monotone transformation and non-interactivity conditions (Sec.~3.2), the interface edges $(v_{\tables}^i, v_s^\ell)$ correspond to feasible pick-and-place transfers from the tabletop abstraction back to stack locations.
Costs need not be a unit action count, and they can be extended to better approximate expected effort (e.g., using stack location distances).

\begin{claim}[Feasible Monotone-Transformed Topple
]
Given an initial arrangement of object pebbles on $G_{SMASH}$ corresponding to the stack rearrangement initial state, any sequence of pick--and--place and topple actions corresponds to a sequence of edge transitions of the pebbles on $G_{SMASH}$. Provided the monotone transformation and non-interactivity conditions (Sec.~3.2) hold, these transitions define candidate task plans in which topple moves objects into the tabletop abstraction and subsequent picks from the tabletop are grounded during execution.
\end{claim}

\subsection{Computing a candidate task plan on the gadget}

We compute a candidate task plan by solving a multi-agent path finding problem on $G_{SMASH}$ with one pebble per object. Following solver strategies for gadget-based rearrangement~\cite{shome2020synchronized}, we formulate a multi-commodity flow with capacity constraints over a time-expanded version of the graph, and solve the resulting MILP with Gurobi. In our setting, the topple transition is modeled explicitly via capacity $\ell$, and we enforce that only one TOPPLE node is active at a time to reflect a single manipulator.


Algorithm~\ref{algo:smash} summarizes the overall procedure. The initial and goal stack states are mapped to pebble configurations on $G_{SMASH}$, after which the MAPF MILP is solved to obtain a flow solution $\Pi_{MAPF}$. We then post-process this solution to remove auxiliary edges and to linearize temporally independent object movements (e.g., across different stacks), yielding the final candidate action sequence $\Pi_{TASK}$.

\begin{algorithm}[t]
	\small
	\DontPrintSemicolon
	\KwIn{Initial stack rearrangement state $(\pose^1_s \cdots \pose^\ell_s )$, \\Target stack rearrangement state $(\pose^1_g \cdots \pose^\ell_g )$}
	\KwOut{Task plan $(\action_1 \cdots \action_h)$}

    $G_{SMASH} \gets ${\sc ConstructGadget}$()$\\
    $p_s \gets ${\sc GetPebbleState}$(\pose^1_s \cdots \pose^\ell_s )$\\
    $p_g \gets ${\sc GetPebbleState}$(\pose^1_g \cdots \pose^\ell_g )$\\
    $\Pi_{MAPF} \gets ${\sc MAPF}$(G_{SMASH}, p_s, p_g)$\\
    $\Pi_{TASK} \gets ${\sc GetActions}$(\Pi_{MAPF})$\\
	\Return{$\Pi_{TASK}$}
	\caption{{\sc SMASH}}
	\label{algo:smash}
\end{algorithm}

\subsection{SMASH Planning and Execution Implementation}

A SMASH planner optimizes over pick, place, and \emph{topple} actions, and motion planning is then performed. Whenever a future motion depends on an uncertain post--topple landing, the corresponding motion is left partially specified and is tagged as a gap task, following partially grounded TAMP.
The motion layer compiles a continuous sequence of plans that respect the task schedule.

For toppling, the end effector is driven from the current arm state to a pre-smash pose defined along either the $x$ or the $y$ axis of the target stack, with a translational offset that guarantees contact in the intended direction as a goal. This motion is generated while ignoring the collision geometries of the toppled objects, since deliberate contact is intended.

For a move involving a toppled object, the system cannot know the pick pose during planning, and it must ensure that the rest of the schedule remains well-posed. An ungrounded motion is created that takes the arm from its current state to the future place pose and stores the terminal arm state needed to reconnect trajectories later. This creates a \emph{gap task} to be solved at execution time once the true pick pose is known.

Execution consumes this partially grounded sequence. When a topple is executed, the simulator returns the new poses of all toppled blocks, and the planning scene is updated with those values. When a gap task is reached, the method plans from the stored arm state to the revealed pick pose, then from pick to place, and finally from place back to the terminal arm state that the ungrounded motion promised. This bridges the gap without breaking the task schedule.


\section{Experiments and Benchmarks}
\label{sec:experiments}

\newcolumntype{C}{>{\centering\arraybackslash}X}
\newcolumntype{Y}{>{\raggedleft\arraybackslash}X}

\begin{table*}[t]
\centering
\small
\setlength{\tabcolsep}{6pt}
\renewcommand{\arraystretch}{1.15}
\begin{tabular*}{\textwidth}{@{\extracolsep{\fill}} l cc cc cc}
\toprule
& \multicolumn{2}{c}{\textbf{4 Blocks}} & \multicolumn{2}{c}{\textbf{6 Blocks}} & \multicolumn{2}{c}{\textbf{9 Blocks}} \\
\cmidrule(lr){2-3}\cmidrule(lr){4-5}\cmidrule(lr){6-7}
\textbf{Metric} & \textbf{w/ topple} & \textbf{w/o topple} & \textbf{w/ topple} & \textbf{w/o topple} & \textbf{w/ topple} & \textbf{w/o topple} \\
\midrule
Success Rate                  & 1.00 & 1.00 & 1.00 & 1.00 & 1.00 & 1.00 \\
Number of Actions             & $2.20\pm0.41$ & $2.40\pm0.82$ & $2.45\pm0.51$ & $3.70\pm2.27$ & $2.95\pm1.19$ & $5.05\pm3.49$ \\
Initial TP Time {[s]}   & $0.223\pm0.005$ & $0.167\pm0.007$ & $0.661\pm0.036$ & $0.488\pm0.043$ & $2.764\pm0.122$ & $3.901\pm4.060$ \\
Offline MP Time {[s]} & $0.46\pm0.03$ & $0.54\pm0.19$ & $0.48\pm0.05$ & $0.82\pm0.51$ & $0.54\pm0.06$ & $1.15\pm0.78$ \\
Execution {[s]}               & $11.06\pm2.20$ & $12.06\pm4.26$ & $12.50\pm2.88$ & $18.48\pm11.14$ & $14.05\pm4.24$ & $25.51\pm17.60$ \\
Goal Object Depth             & $1.20\pm0.4104$ & $1.20\pm0.4104$ & $1.85\pm1.1367$ & $1.85\pm1.1367$ & $2.55\pm1.5035$ & $2.55\pm1.5035$ \\
\bottomrule

\end{tabular*}
\caption{Single-goal rearrangement results. Values are means $\pm$ standard deviation over 20 random instances per setting.
Within each size, w/ topple means toppling enabled and w/o topple means toppling disabled.
We do not report task-planning time explicitly because task planning is solved by Gurobi with a fixed time budget (30\,s) and 12 auxiliary buffers for all methods, we therefore report the time to first feasible task planning solution. Reported motion planning times are offline computation.
Online motion planning is zero in this benchmark because the goal object is not toppled, hence no post-topple pick depends on an unknown pose.}
\label{tab:single}
\end{table*}

\begin{table*}[t]
\centering
\small
\setlength{\tabcolsep}{6pt}
\renewcommand{\arraystretch}{1.15}
\begin{tabular*}{\textwidth}{@{\extracolsep{\fill}} l cc cc cc}
\toprule
& \multicolumn{2}{c}{\textbf{4 Blocks}} & \multicolumn{2}{c}{\textbf{6 Blocks}} & \multicolumn{2}{c}{\textbf{9 Blocks}} \\
\cmidrule(lr){2-3}\cmidrule(lr){4-5}\cmidrule(lr){6-7}
\textbf{Metric} & \textbf{w/ topple} & \textbf{w/o topple} & \textbf{w/ topple} & \textbf{w/o topple} & \textbf{w/ topple} & \textbf{w/o topple} \\
\midrule
Success Rate                  & 1.00 & 1.00 & 0.95 & 1.00 & 1.00 & 1.00 \\
Number of Actions             & $8.05\pm2.01$  & $9.90\pm3.08$  & $12.90\pm1.45$ & $16.80\pm2.55$ & $20.85\pm4.21$ & $26.00\pm6.74$ \\
Initial TP Time {[s]}   & $0.172\pm0.005$ & $0.132\pm0.005$ & $0.530\pm0.021$ & $0.528\pm0.323$ & $3.139\pm2.305$ & $3.389\pm1.840$ \\
Offline MP Time {[s]} & $1.12\pm0.26$  & $2.23\pm0.70$  & $1.60\pm0.32$ & $3.79\pm0.59$ & $2.97\pm0.58$ & $5.97\pm1.61$ \\
Online MP Time {[s]}  & $0.77\pm0.33$  & $0.00\pm0.00$  & $2.00\pm3.32$ & $0.00\pm0.00$ & $1.71\pm0.74$ & $0.00\pm0.00$ \\
Execution {[s]}               & $43.38\pm12.38$ & $49.98\pm16.03$ & $70.11\pm10.12$ & $84.20\pm12.97$ & $113.58\pm25.72$ & $129.28\pm34.81$ \\
\bottomrule
\end{tabular*}
\caption{Multi-goal rearrangement results. Values are means $\pm$ standard deviation over 20 random instances per setting.
Within each size, w/ topple means toppling enabled and w/o topple means toppling disabled.
The same resource setup is used (30 s budget, 12 buffers).
Success is defined as: the task optimizer returns a feasible schedule within 30\,s and all required motions 
execute successfully. Failures with toppling are mainly due to post-topple objects landing beyond the robot’s reachable workspace or in poses that violate grasp/IK constraints during online gap bridging.}
\label{tab:multi}
\end{table*}

\begin{figure*}[t!]
    \centering
    \includegraphics[width=0.99\linewidth]{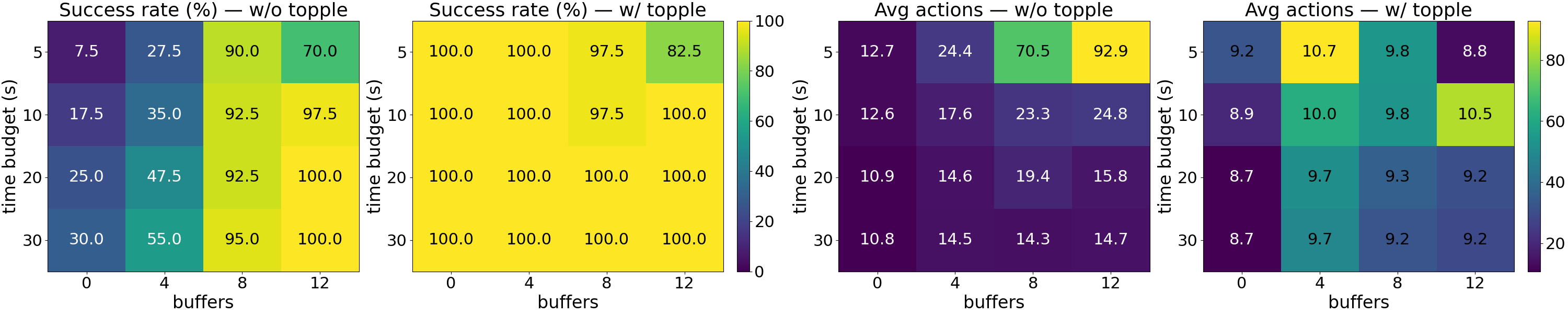}
    \caption{Task planning ablation for 9-block instances from the multi-goal rearrangement benchmark. Left: success rate (\%) as a function of task-planning time budget and buffer count. Right: average number of actions over successful runs. Each heatmap compares no-topple vs topple under identical settings.}
    \label{fig:horizon-heatmap}
    \vspace{-0.15in}
\end{figure*}

\begin{figure}[ht]
    \centering
    \includegraphics[trim=10cm 0 10cm 0,clip,width=0.29\linewidth]{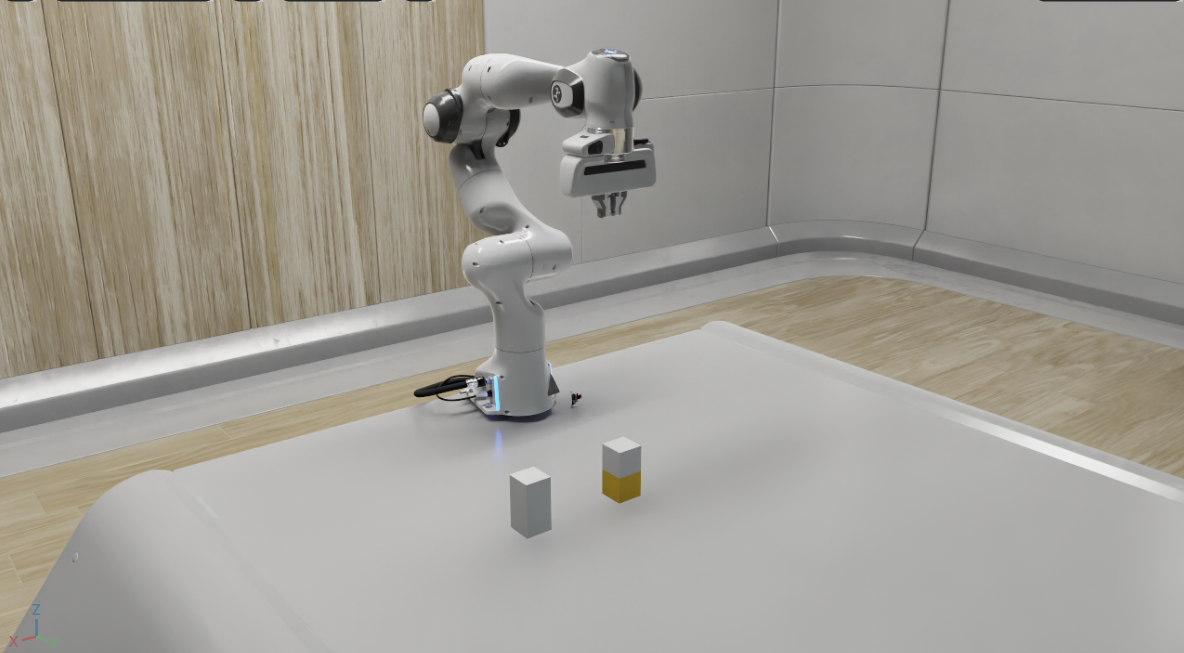}
    \includegraphics[trim=10cm 0 10cm 0,clip,width=0.29\linewidth]{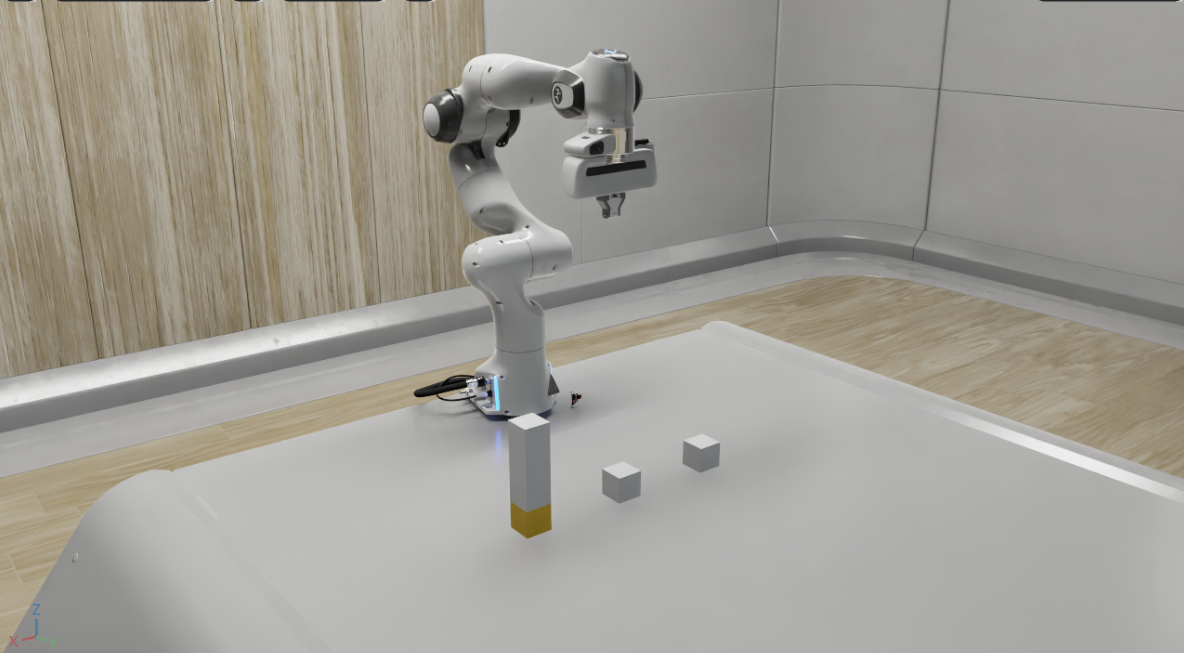}
    \includegraphics[trim=10cm 0 10cm 0,clip,width=0.29\linewidth]{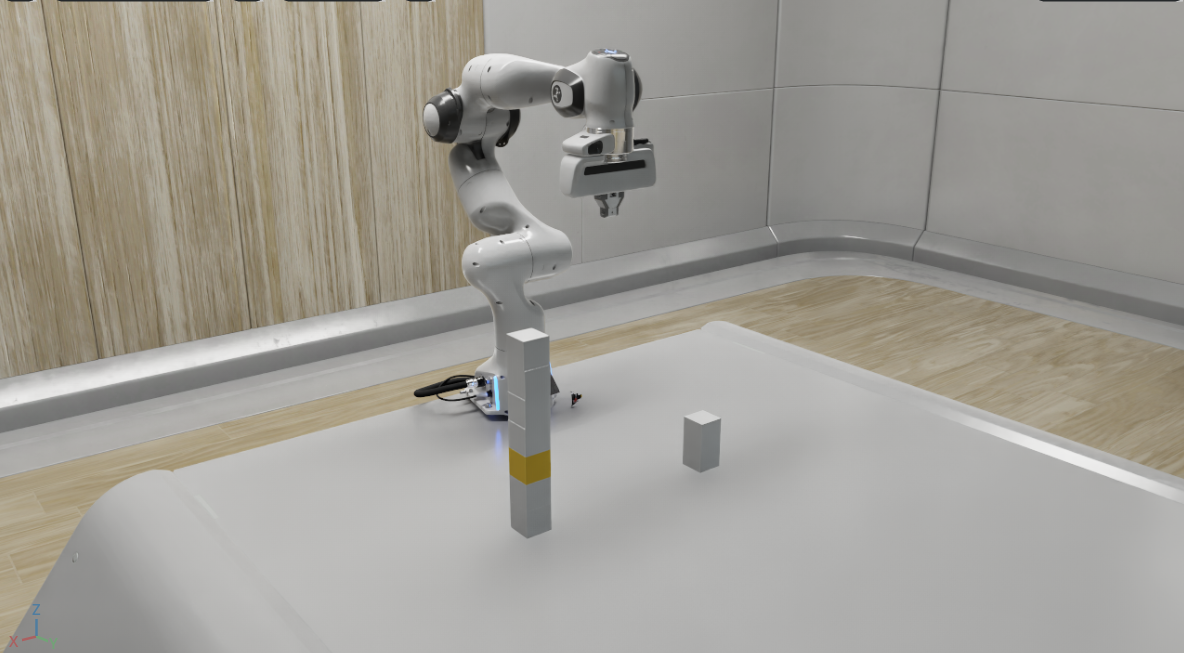}

    \vspace{0.05in}

    \centering
    \includegraphics[trim=10cm 0 10cm 0,clip,width=0.29\linewidth]{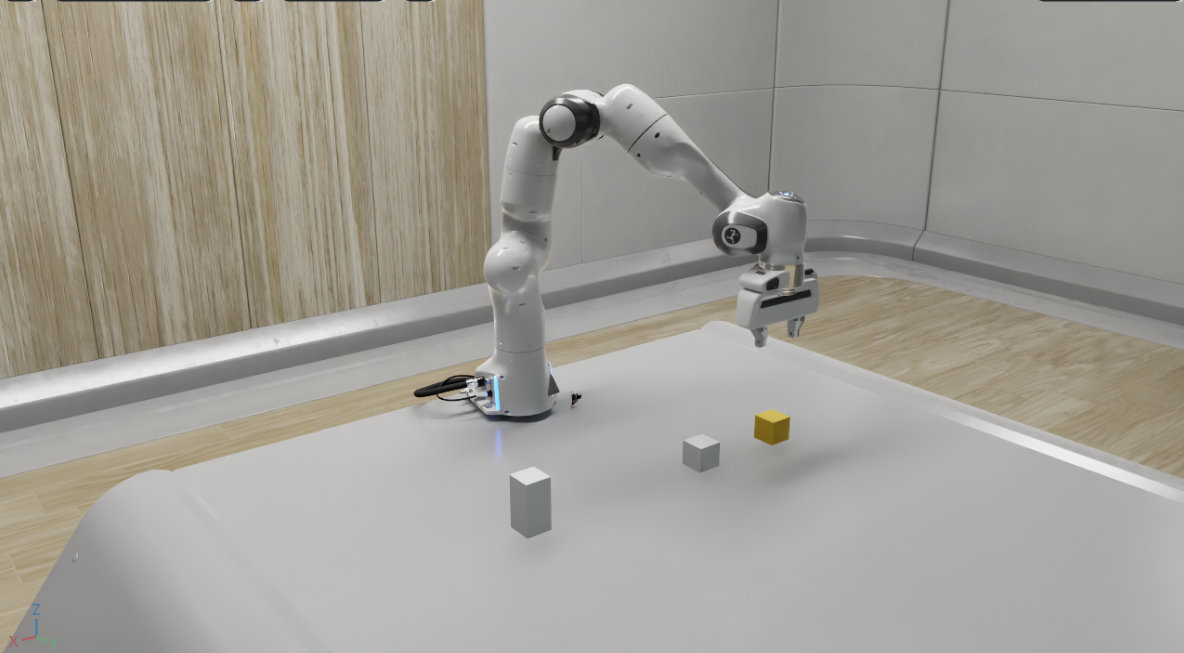}
    \includegraphics[trim=10cm 0 10cm 0,clip,width=0.29\linewidth]{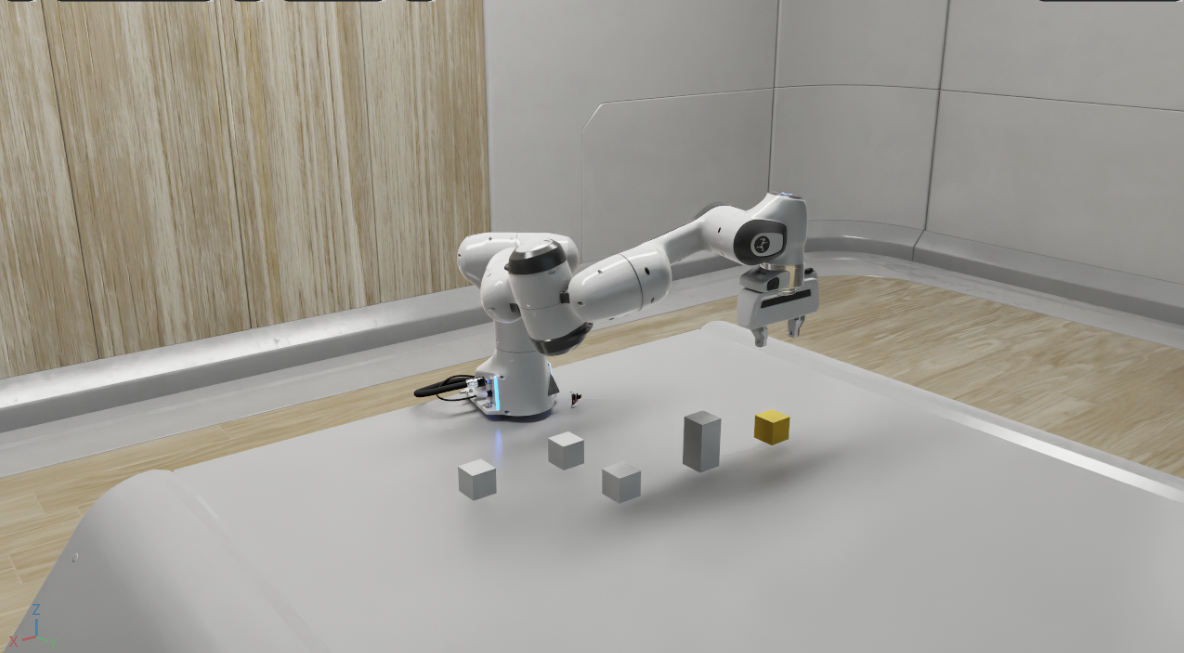}
    \includegraphics[trim=10cm 0 10cm 0,clip,width=0.29\linewidth]{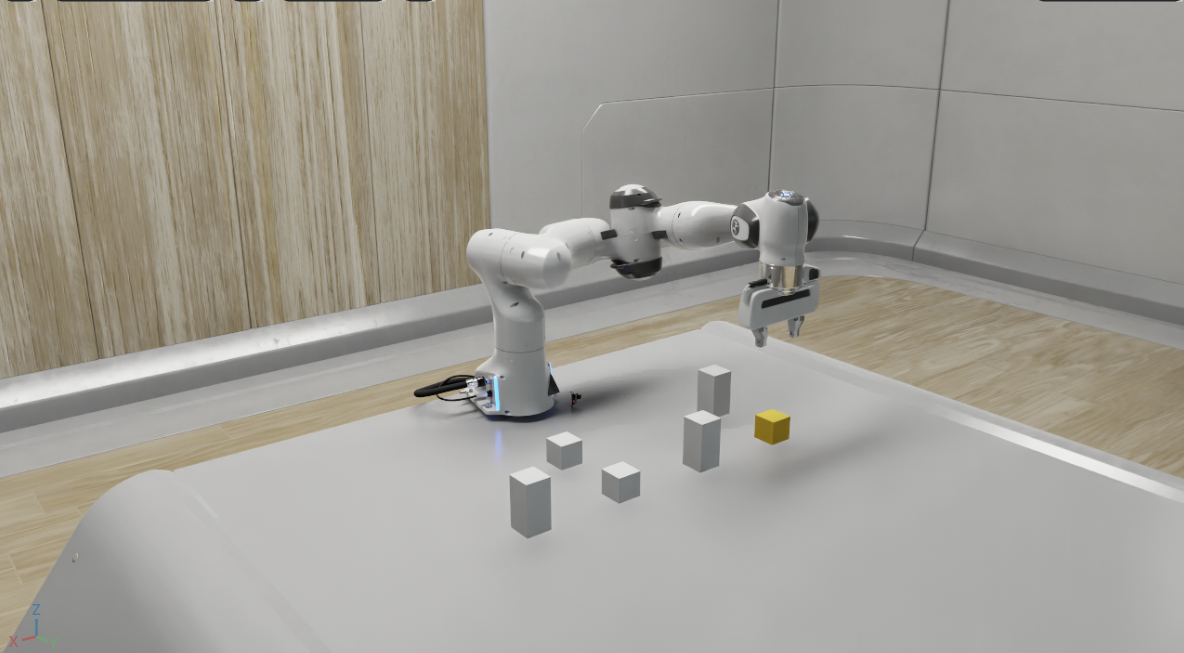}

    \vspace{0.05in}
    
    \includegraphics[trim=10cm 0 10cm 0,clip,width=0.29\linewidth]{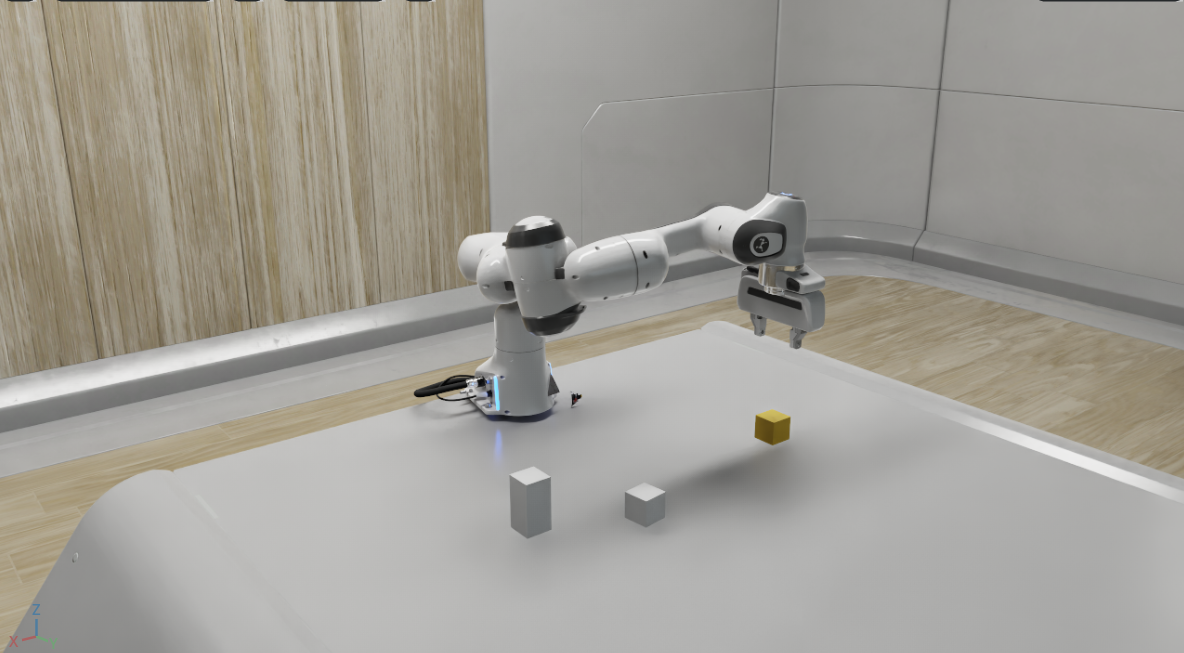}
    \includegraphics[trim=10cm 0 10cm 0,clip,width=0.29\linewidth]{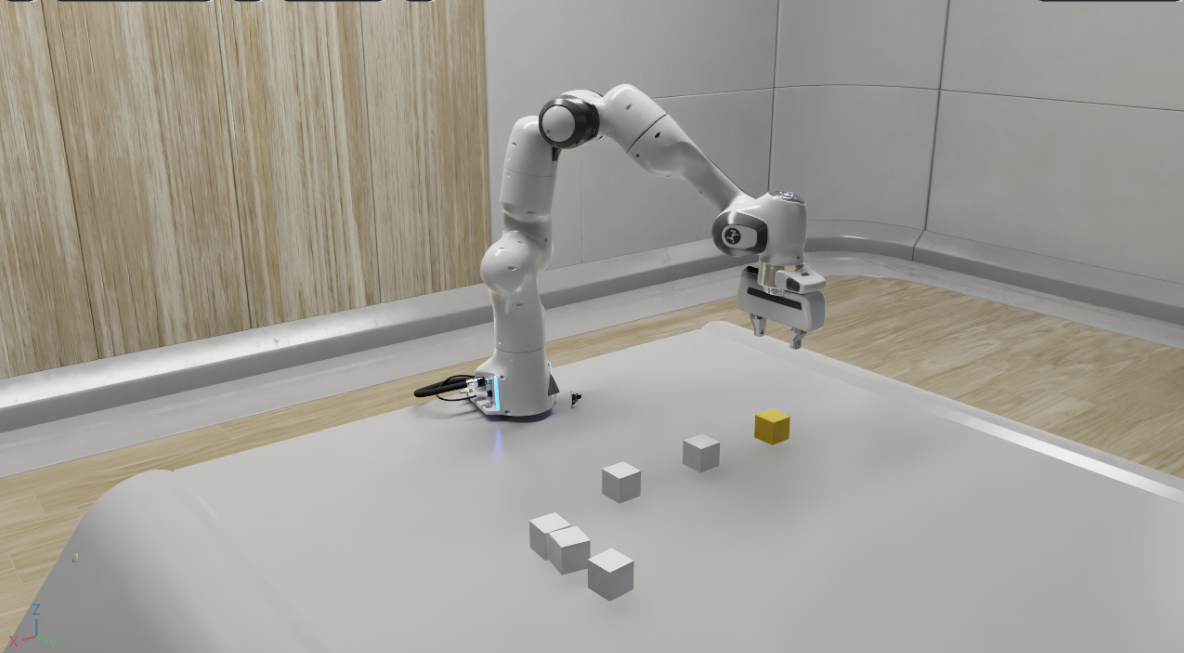}
    \includegraphics[trim=10cm 0 10cm 0,clip,width=0.29\linewidth]{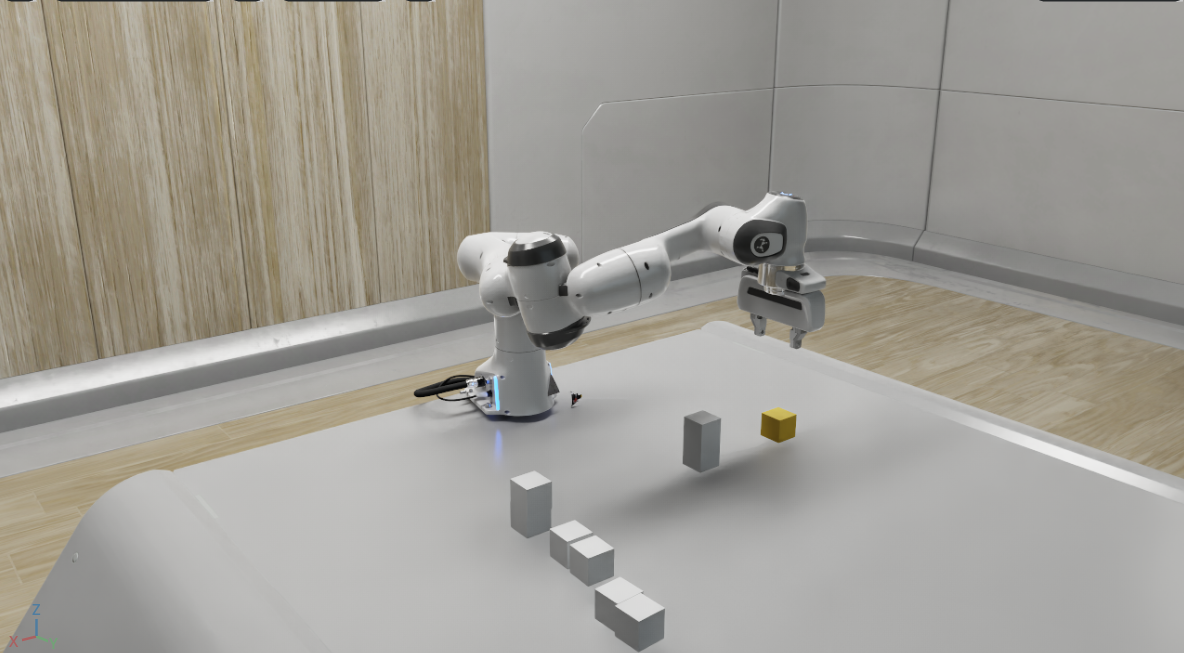}
    
    \setlength{\tabcolsep}{10pt}
    \newcolumntype{Y}{>{\centering}X}
    \begin{tabularx}{0.87\linewidth}{Y Y Y Y}
        \smaller \textbf{4 blocks} & \smaller \textbf{       6 blocks} & \smaller \textbf{          9 blocks}
    \end{tabularx}
    \caption{The figure shows the start and goal arrangements of the single-goal benchmark problem instances for 4, 6, and 9 blocks. Top row: start arrangement. Middle row: goal arrangement after \textit{w/o topple} execution. Bottom row: goal state of after \textit{w/ topple} execution.}
    \vspace{-0.2in}
    \label{img:single}
\end{figure}

We evaluate the SMASH planner on stacked-block manipulation with and without toppling. SMASH produces a \emph{partially grounded} task-and-motion plan in which actions whose feasibility depends on post-topple poses are deferred and bridged online at execution time~\cite{pan2024task}. All experiments are run in Isaac Sim with a Franka Panda 7-DoF robotic arm and a parallel gripper. Experiments run on an AMD Ryzen 9 9950X with $128$\,GB RAM, using Gurobi~\cite{gurobi} for task optimization and a standard ROS2~\cite{scirobotics.abm6074}/MoveIt2~\cite{coleman2014reducing}/OMPL~\cite{sucan2012open} stack with RRTConnect~\cite{844730} for motion planning, backed by the Nvidia IsaacSim and PhysX engine for dynamics.

\noindent\textbf{Tasks and scenes.}
We consider two problem settings: (i) \emph{single-goal rearrangement}, where one designated block must be moved to a target location, and (ii) \emph{multi-goal rearrangement}, where all blocks must reach designated goal locations. For each benchmark we test stacks of \{4, 6, 9\} blocks, and generate $20$ random instances per setting.

\noindent\textbf{Metrics.}
We compare a task planner using the proposed SMASH gadget (``w/ topple'') with a variant only using pick and place actions (``w/o topple'').
We report success rate, number of actions, initial task planning (TP) time, \emph{offline} motion planning (MP) time, \emph{online} motion-planning (MP) time. Note that only interacting with toppled objects introduces the need to motion plan during execution since these object poses cannot be grounded offline (before execution). In constrast pick and place (w/o topple) performs open-loop execution of the offline-computed plan. The execution time on IsaacSim is also reported.
Task planning is solved by optimizing the MAPF problem in Algo~\ref{algo:smash} using Gurobi with a fixed time budget of 30\,s per instance. Gurobi has anytime behavior and we 
report the time to the \emph{first feasible} task plan. Success requires a feasible task plan within the budget and successful execution of all required motions in the simulator.

\subsection{Benchmark: Single-Goal Rearrangement}
\label{subsec:single-goal}
This benchmark (Fig.~\ref{img:single}) randomizes stacks with a single target object at some \textit{goal object depth} in the stack with the goal always defined to be a predefined location.
Table~\ref{tab:single} reports single-goal results. Both methods succeed on all instances, but enabling toppling reduces action count and reduces execution time, with larger gains as the goal becomes more deeply buried in the stack. 
Toppling clears multiple constraints in one step instead of multiple intermediate pick-and-place actions.
As \emph{goal object depth} increases, efficiency benefits of toppling get pronounced.

\subsection{Benchmark: Multi-Goal Rearrangement}
\label{subsec:multi-goal}

The randomized multi-goal benchmark samples stacks with different numbers of objects alongside randomized goal stacks. 
Table~\ref{tab:multi} reports multi-goal results under a fixed task-planning budget of 30\,s and 12 additional object locations as buffer for each run for both methods. Under these resources, both variants achieve high success rates. Enabling toppling consistently reduces the number of discrete manipulation actions and correspondingly reduces 
overall execution time in contrast to \textit{w/o topple}.
Online motion-planning time is needed for plans that interact with toppled objects, as some pick poses can only be determined after the physical outcome of the topple is observed and the state updated. Despite the online bridging cost, the overall planning and execution profiles maintain efficiency with toppling.
A failure is observed in one of the (6 obj) runs with toppling (Table~\ref{tab:multi}).
After a topple executes, some objects do not lie in the robot’s feasibly reachable workspace.
In such cases, the required gap-bridging motion cannot be generated online during execution, leading to failure despite the existence of a candidate task-level schedule. 
\begin{wrapfigure}[16]{r}{0.4\linewidth}
    \centering
    \vspace{-0.1in}
    \includegraphics[width=0.95\linewidth]{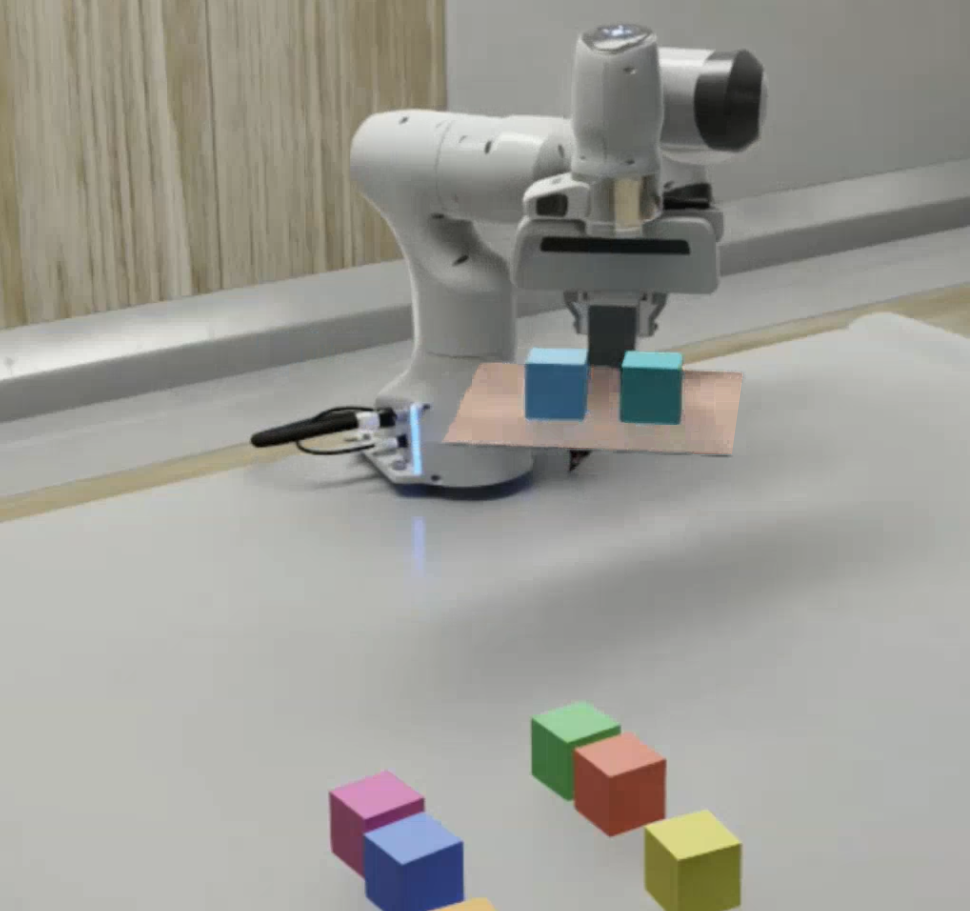}
    \includegraphics[width=0.95\linewidth]{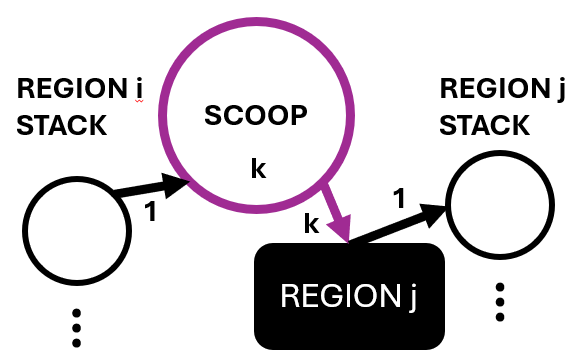}
    \caption{A scoop action (above) can use similar gadgets (below). }
    \vspace{-0.4in}
    \label{fig:scoop}
\end{wrapfigure}
This failure introduces an interesting trade-off for the efficiency benefits obtained from toppling -- 
toppling remains a dynamic, uncertain action whose exact grounding is resolved only at execution time. This motivates future investigation into characterizing robust execution strategies.
Fig.~\ref{img:multi} shows representative multi-goal instances. In Fig.~\ref{img:multi}, both methods reach the same final arrangement (multi-goal), and the difference is in the action sequence: toppling collapses the stacks earlier, reducing intermediate relocations.

\subsection{Task Planning Ablation}
\label{subsec:horizon}

To understand when task planning becomes compute- and buffer-limited, we evaluate the \emph{task planner alone} on 9-block instance of the multi-goal rearrangement benchmark over different combinations of (i) the Gurobi time budget and (ii) the number of available \emph{auxiliary} buffer stack locations. 
Fig.~\ref{fig:horizon-heatmap} summarizes the ablation.
Two observations stand out. First, toppling substantially relaxes the feasible planning horizon under tight resources: with \emph{no auxiliary buffers} ($B{=}0$), topple achieves $\ge 90\%$ success across all tested time budgets, whereas no-topple exhibits low success at $B{=}0$ even as the budget increases. Intuitively, toppling acts simultaneously on multiple objects in a single step of the underlying planning search tree, thereby short-cutting the exploration of the multiple pick and place actions necessary across deeper horizons. Second, \textit{w/o topple} requires substantially more buffer capacity to reliably find feasible schedules under a fixed budget: in our sweep, the smallest auxiliary buffer count that reaches $\ge 90\%$ success for no-topple is $B{=}8$, while topple reaches this threshold already at $B{=}0$. Among successful runs, topple also yields consistently shorter schedules. At matched buffers, the average action count is reduced by roughly $1.5\times$--$7\times$ depending on the budget (Fig.~\ref{fig:horizon-heatmap}, right).
We use $12$ buffers and a 30\,s budget in the main benchmarks where both methods hit perfect success and \textit{w/ topple} saturates solution improvement.

\begin{figure}[t]
    \centering
    \includegraphics[trim=10cm 0 10cm 0,clip,width=0.29\linewidth]{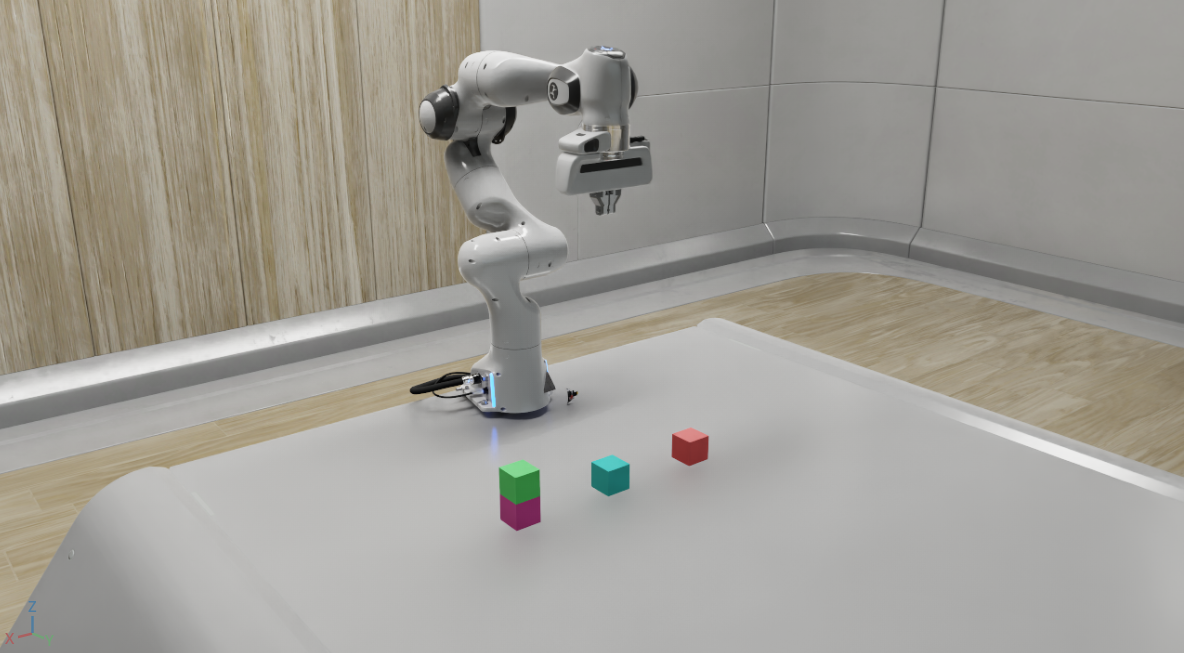}
    \includegraphics[trim=10cm 0 10cm 0,clip,width=0.29\linewidth]{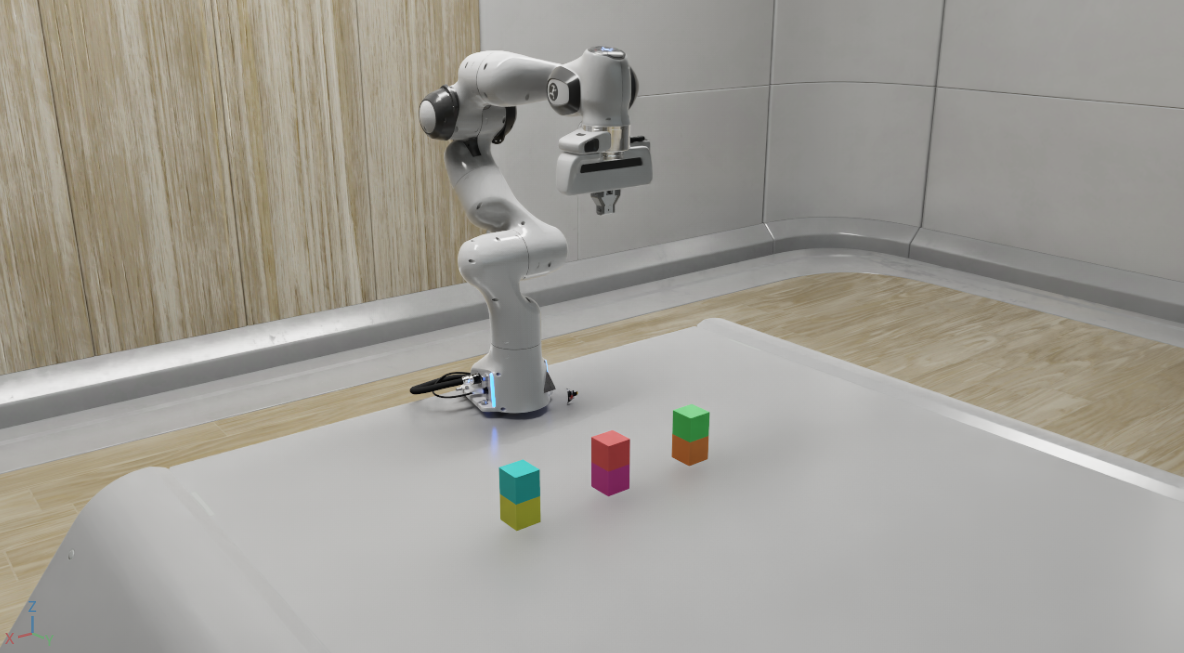}
    \includegraphics[trim=10cm 0 10cm 0,clip,width=0.29\linewidth]{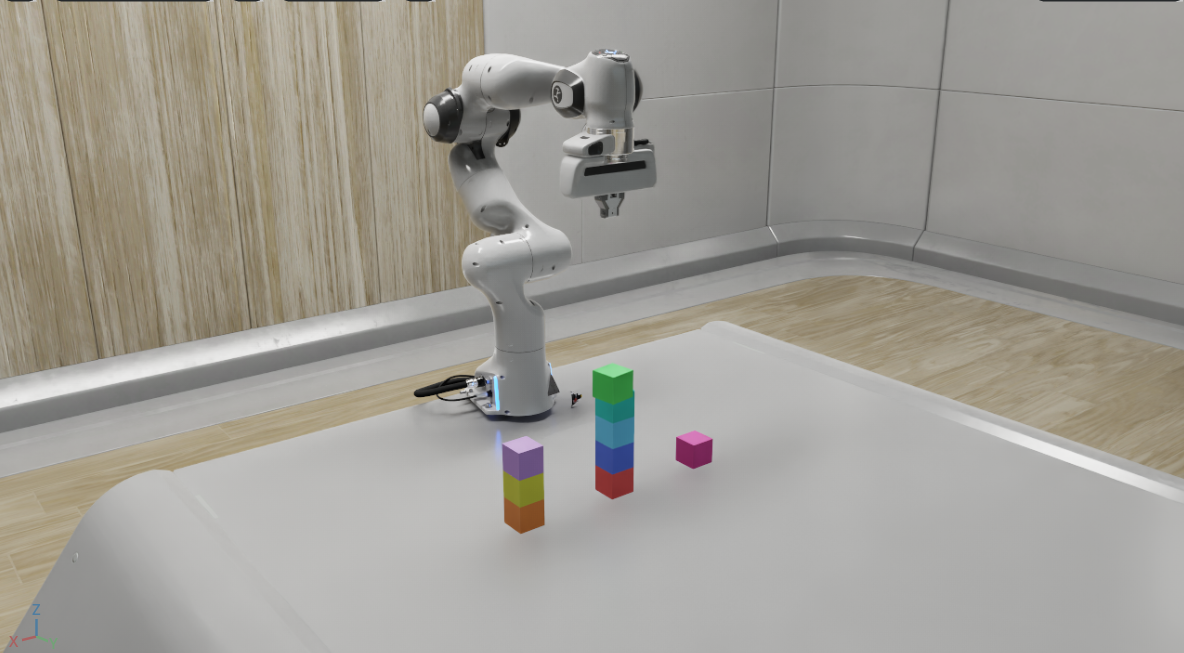}

    \vspace{0.05in}

    \centering
    \includegraphics[trim=10cm 0 10cm 0,clip,width=0.29\linewidth]{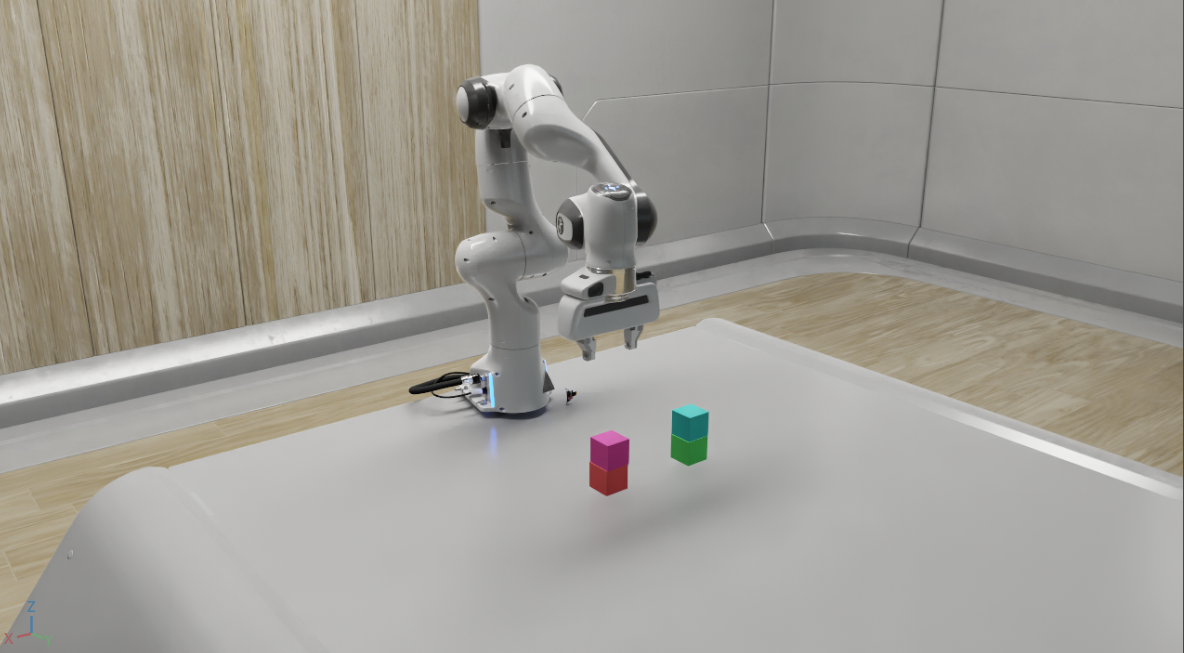}
    \includegraphics[trim=10cm 0 10cm 0,clip,width=0.29\linewidth]{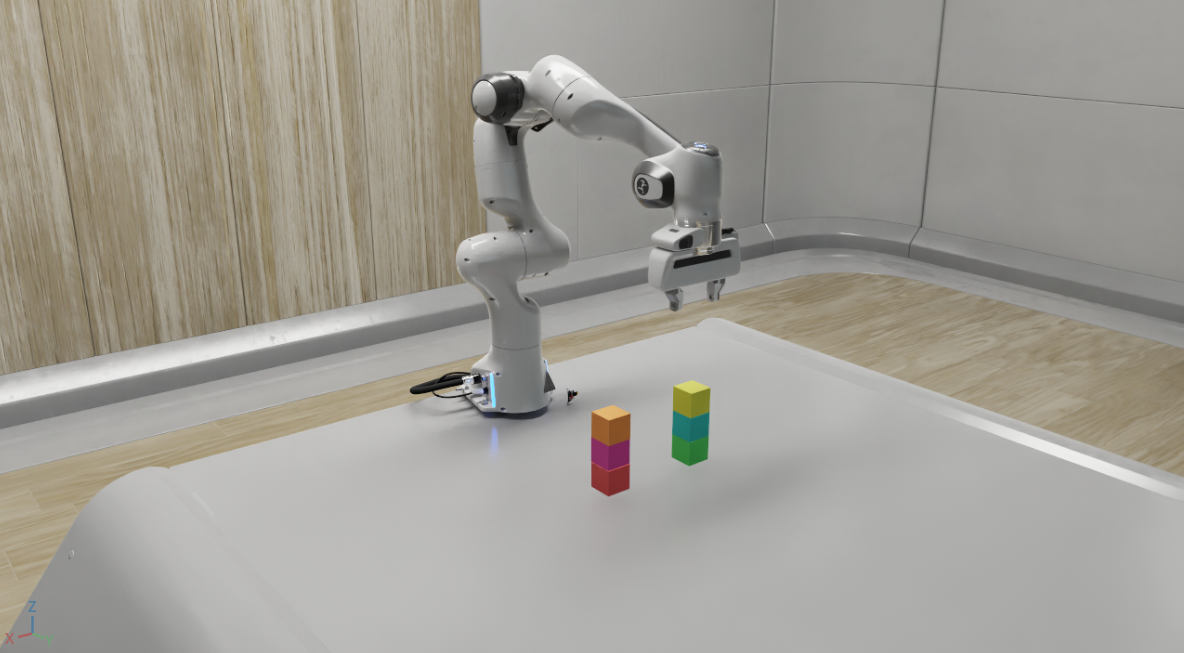}
    \includegraphics[trim=10cm 0 10cm 0,clip,width=0.29\linewidth]{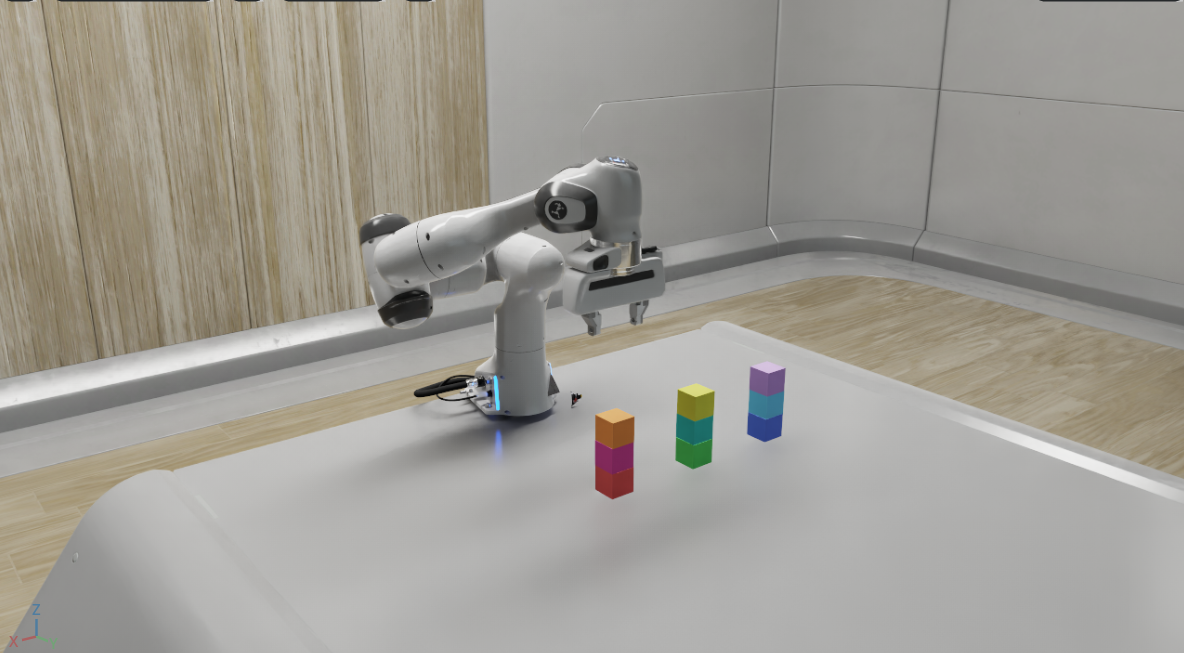}

    \vspace{0.05in}

    \setlength{\tabcolsep}{10pt}
    \newcolumntype{Y}{>{\centering}X}
    \begin{tabularx}{0.87\linewidth}{Y Y Y Y}
        \smaller \textbf{4 blocks} & \smaller \textbf{       6 blocks} & \smaller \textbf{          9 blocks}
    \end{tabularx}
    \caption{The figure shows the start and goal arrangements of the multi-goal benchmark problem instances 
    for 4, 6, and 9 blocks. Top row: start arrangement. 
    Bottom row: goal arrangement of problem instance for both \textit{w/o topple} and \textit{w/ topple} executions.
    }
    \label{img:multi}
    \vspace{-0.2in}
\end{figure}


\subsection{Demonstration: Scoop}
The proposed abstraction is not limited to toppling. It supports rich simultaneous manipulation primitives modeled as capacity-constrained aggregate transitions. To demonstrate this generality, we introduce a \emph{scoop} action with a tray-like object that can transport multiple objects simultaneously.

In the demonstration scenario (Fig.~\ref{fig:scoop}), two stacks of four blocks are located on the left side of the table, alongside the scoop. The goal is to relocate the two bottom blocks (one from each stack) to designated positions on the right side of the workspace. Without toppling or scooping, the planner must first remove the six upper blocks via pick-and-place before transferring the two bottom blocks individually across the workspace.
When topple and scoop are enabled, the planner first topples both stacks, exposing the bottom blocks. It then transfers the two target blocks onto the scoop using local pick-and-place operations. Finally, the scoop itself—carrying both blocks—is transported in a single motion to the right side of the workspace, after which the blocks are placed at their designated locations.
This interleaving of topple, scoop, and pick-and-place compresses the manipulation horizon and reduces long-distance transfers. In IsaacSim simulation, this strategy yields a $14\%$ reduction in execution time compared to a pick-and-place-only solution. This demonstration highlights that the similar abstractions can support a broader class of actions beyond toppling.


\section{Discussion}
This work studies tabletop stack rearrangement when a non-prehensile \emph{topple} action is available, with the goal of exploiting simultaneous multi-object interactions while acknowledging that their outcomes cannot be fully grounded at plan time. We introduced a task-level abstraction, the SMASH gadget, that augments prior stack-rearrangement gadgets with capacity-aware transitions representing topple, enabling automatic interleaving of pick--and--place and topple in candidate task plans. We then integrated this task plan with partially grounded task-and-motion execution. Across both benchmarks, enabling toppling consistently reduces action count and decreases both planning and execution time, reflecting shorter and less constrained schedules. 

Several extensions could improve robustness while preserving the core abstraction. A simple task-level mechanism is to restrict how many objects may be toppled in one action, which biases planning toward smaller topples that reduce landing dispersion. In addition, execution feedback can be used to adapt the task layer: topple patterns (e.g., stack index and topple height) that systematically yield unreachable landings can be penalized or pruned in subsequent planning attempts. 
Future studies can explore incorporating additional affordance constraints and uncertainty in robust online primitives. The motivating scoop demonstration shows the promise of leveraging the use of graphical abstractions as the one proposed for toppling to further explore richer manipulation spanning non-prehensile actions, aggregating, decomposing, cutting actions, or tool usage. The current work is a necessary first step that holds promise of richer dynamic interactions empowering richer manipulation applications.

\bibliography{refs}

@article{sucan2012open,
  title={The open motion planning library},
  author={Sucan, Ioan A and Moll, Mark and Kavraki, Lydia E},
  journal={IEEE Robotics \& Automation Magazine},
  volume={19},
  number={4},
  pages={72--82},
  year={2012},
  publisher={IEEE}
}

@article{coleman2014reducing,
  title={Reducing the barrier to entry of complex robotic software: a {{MoveIT}}! case study},
  author={Coleman, David and Sucan, Ioan and Chitta, Sachin and Correll, Nikolaus},
  journal={arXiv preprint arXiv:1404.3785},
  year={2014}
}

@misc{gurobi,
  author = {{Gurobi Optimization, LLC}},
  title = {{Gurobi Optimizer Reference Manual}},
  year = 2023
}

@article{
    scirobotics.abm6074,
    author = {Steven Macenski  and Tully Foote  and Brian Gerkey  and Chris Lalancette  and William Woodall },
    title = {Robot Operating System 2: Design, architecture, and uses in the wild},
    journal = {Science Robotics},
    volume = {7},
    number = {66},
    pages = {eabm6074},
    year = {2022},
    doi = {10.1126/scirobotics.abm6074}
}

@InProceedings{CLIP,
  title = 	 {Learning Transferable Visual Models From Natural Language Supervision},
  author =       {Radford, Alec and Kim, Jong Wook and Hallacy, Chris and Ramesh, Aditya and Goh, Gabriel and Agarwal, Sandhini and Sastry, Girish and Askell, Amanda and Mishkin, Pamela and Clark, Jack and Krueger, Gretchen and Sutskever, Ilya},
  booktitle ={Proceedings of the International Conference on Machine Learning},
  pages = 	 {8748--8763},
  year = 	 {2021},
  editor = 	 {Meila, Marina and Zhang, Tong},
  volume = 	 {139},
  series = 	 {Proceedings of Machine Learning Research},
  publisher =    {PMLR}
}

@inproceedings{Kaelbling:2011gb,
	Author = {Kaelbling, L. P. and Lozano-P\'{e}rez, T.},
	Booktitle = {ICRA},
	Title = {{Hierarchical Task and Motion Planning in the Now}},
	Year = {2011}}

@inproceedings{stilman2007manipulation,
  title={Manipulation planning among movable obstacles},
  author={Stilman, Mike and Schamburek, Jan-Ullrich and Kuffner, James and Asfour, Tamim},
  booktitle={ICRA},
  year={2007}
}

@inproceedings{ota2004rearrangement,
  title={Rearrangement of multiple movable objects-integration of global and local planning methodology},
  author={Ota, Jun},
  booktitle={ICRA},
  volume={2},
  year={2004}
}

@article{han2017complexity,
  title={Complexity Results and Fast Methods for Optimal Tabletop Rearrangement with Overhand Grasps},
  author={Han, Shuai D and Stiffler, Nicholas M and Krontiris, Athanasios and Bekris, Kostas and Yu, Jingjin},
  journal={arXiv:1711.07369},
  year={2017}
}

@article{garrett2018ffrob,
  title={FFRob: Leveraging symbolic planning for efficient task and motion planning},
  author={Garrett, Caelan Reed and Lozano-Perez, Tomas and Kaelbling, Leslie Pack},
  journal={IJRR},
  volume={37},
  number={1},
  pages={104--136},
  year={2018},
  publisher={SAGE Publications Sage UK: London, England}
}

@article{akbari2018combined,
	Author = {Akbari, Aliakbar and Lagriffoul, Fabien and Rosell, Jan},
	Journal = {Autonomous Robots},
	Pages = {1--16},
	Publisher = {Springer},
	Title = {Combined heuristic task and motion planning for bi-manual robots},
	Year = {2018}}

@inproceedings{shome2019towards,
	Author = {Shome, Rahul and Tang, Wei N and Song, Changkyu and Mitash, Chaitanya and Kourtev, Hristiyan and Yu, Jingjin and Boularias, Abdeslam and Bekris, Kostas E},
	Booktitle = {IEEE International Conference on Robotics and Automation (ICRA)},
	Title = {Towards Robust Product Packing with a Minimalistic End-Effector},
	Year = {2019}}

@article{han2018efficient,
  title={Efficient, high-quality stack rearrangement},
  author={Han, Shuai D and Stiffler, Nicholas M and Bekris, Kostas E and Yu, Jingjin},
  journal={IEEE RAL},
  volume={3},
  number={3},
  pages={1608--1615},
  year={2018},
  publisher={IEEE}
}

@article{dantam2018incremental,
    title = {{An incremental constraint-based framework for task and motion planning}},
    author = {Dantam, Neil T and Kingston, Zachary K and Chaudhuri, Swarat and Kavraki, Lydia E},
    journal = {{The International Journal of Robotics Research}},
    volume = {37},
    number = {10},
    pages = {1134--1151},
    year = {2018},
    publisher = {SAGE Publications Sage UK: London, England}
}

@INPROCEEDINGS{curtis2022M0M,
    author = {Curtis, Aidan and Fang, Xiaolin and Kaelbling, Leslie Pack and Lozano-Pérez, Tomás and Garrett, Caelan Reed},
    booktitle = {{2022 International Conference on Robotics and Automation (ICRA)}},
    title = {{Long-horizon manipulation of unknown objects via task and motion planning with estimated Affordances}},
    year = {2022},
    volume = {},
    number = {},
    pages = {1940-1946},
    doi = {10.1109/ICRA46639.2022.9812057}
}

@inproceedings{pan2022failure,
    title = {{Failure is an option: task and motion planning with failing executions}},
    author = {Pan, Tianyang and Wells, Andrew M and Shome, Rahul and Kavraki, Lydia E},
    booktitle = {{2022 International Conference on Robotics and Automation (ICRA)}},
    pages = {1947--1953},
    year = {2022},
    organization = {IEEE}
}

@inproceedings{shah2020anytime,
    title = {{Anytime integrated task and motion policies for stochastic environments}},
    author = {Shah, Naman and Vasudevan, Deepak Kala and Kumar, Kislay and Kamojjhala, Pranav and Srivastava, Siddharth},
    booktitle = {{2020 IEEE International Conference on Robotics and Automation (ICRA)}},
    pages = {9285--9291},
    year = {2020},
    organization = {IEEE}
}

@article{garrett2021review,
    author = {Garrett, Caelan Reed and Chitnis, Rohan and Holladay, Rachel and Kim, Beomjoon and Silver, Tom and Kaelbling, Leslie Pack and Lozano-P\'{e}rez, Tom\'{a}s},
    title = {{Integrated task and motion planning}},
    journal = {{Annual Review of Control, Robotics, and Autonomous Systems}},
    volume = {4},
    number = {1},
    pages = {265-293},
    year = {2021},
    doi = {10.1146/annurev-control-091420-084139}
}

@article{pan2024task,
  title={Task and motion planning for execution in the real},
  author={Pan, Tianyang and Shome, Rahul and Kavraki, Lydia E},
  journal={IEEE Transactions on Robotics},
  volume={40},
  pages={3356--3371},
  year={2024},
  publisher={IEEE}
}

@inproceedings{shome2020synchronized,
  title={Synchronized multi-arm rearrangement guided by mode graphs with capacity constraints},
  author={Shome, Rahul and Bekris, Kostas E},
  booktitle={International Workshop on the Algorithmic Foundations of Robotics},
  pages={243--260},
  year={2020},
  organization={Springer}
}

@inproceedings{vieira2022persistent,
  title={Persistent homology for effective non-prehensile manipulation},
  author={Vieira, Ewerton R and Nakhimovich, Daniel and Gao, Kai and Wang, Rui and Yu, Jingjin and Bekris, Kostas E},
  booktitle={2022 International Conference on Robotics and Automation (ICRA)},
  pages={1918--1924},
  year={2022},
  organization={IEEE}
}

@article{dogar2011framework,
  title={A framework for push-grasping in clutter},
  author={Dogar, Mehmet and Srinivasa, Siddhartha},
  journal={Robotics: Science and systems VII},
  volume={1},
  pages={65--72},
  year={2011},
  publisher={MIT Press Cambridge, MA}
}

@inproceedings{huang2019large,
  title={Large-scale multi-object rearrangement},
  author={Huang, Eric and Jia, Zhenzhong and Mason, Matthew T},
  booktitle={2019 international conference on robotics and automation (ICRA)},
  pages={211--218},
  year={2019},
  organization={IEEE}
}

@inproceedings{pan2021decision,
  title={Decision making in joint push-grasp action space for large-scale object sorting},
  author={Pan, Zherong and Hauser, Kris},
  booktitle={2021 IEEE international conference on robotics and automation (ICRA)},
  pages={6199--6205},
  year={2021},
  organization={IEEE}
}

@inproceedings{lynch1999toppling,
  title={Toppling manipulation},
  author={Lynch, Kevin M},
  booktitle={Proceedings 1999 IEEE International Conference on Robotics and Automation (Cat. No. 99CH36288C)},
  volume={4},
  pages={2551--2557},
  year={1999},
  organization={IEEE}
}

@article{chi2023diffusion,
  title={Diffusion policy: Visuomotor policy learning via action diffusion},
  author={Chi, Cheng and Xu, Zhenjia and Feng, Siyuan and Cousineau, Eric and Du, Yilun and Burchfiel, Benjamin and Tedrake, Russ and Song, Shuran},
  journal={The International Journal of Robotics Research},
  pages={02783649241273668},
  year={2023},
  publisher={SAGE Publications Sage UK: London, England}
}

@inproceedings{vieira2024morals,
  title={Morals: Analysis of high-dimensional robot controllers via topological tools in a latent space},
  author={Vieira, Ewerton R and Sivaramakrishnan, Aravind and Tangirala, Sumanth and Granados, Edgar and Mischaikow, Konstantin and Bekris, Kostas E},
  booktitle={2024 IEEE International Conference on Robotics and Automation (ICRA)},
  pages={27--33},
  year={2024},
  organization={IEEE}
}

@INPROCEEDINGS{844730,
  author={Kuffner, J.J. and LaValle, S.M.},
  booktitle={Proceedings 2000 ICRA. Millennium Conference. IEEE International Conference on Robotics and Automation. Symposia Proceedings (Cat. No.00CH37065)}, 
  title={RRT-connect: An efficient approach to single-query path planning}, 
  year={2000},
  volume={2},
  number={},
  pages={995-1001 vol.2},
  doi={10.1109/ROBOT.2000.844730}}

\end{document}